\documentclass[sigconf]{acmart}

\usepackage{CJKutf8}
\usepackage{amsthm} 

\usepackage{subfigure}
\usepackage{graphicx}
\usepackage{multirow}%
\usepackage{float}
\usepackage{booktabs}
\usepackage{subcaption}
\usepackage[inline]{enumitem}
\usepackage{acronym}
\usepackage{makecell}

\usepackage{CJKutf8}

\AtBeginDocument{%
  \providecommand\BibTeX{{%
    \normalfont B\kern-0.5em{\scshape i\kern-0.25em b}\kern-0.8em\TeX}}}

\acrodef{RL}{Reinforcement Learning}
\acrodef{KG}{Knowledge Graph}
\acrodef{KGE}{Knowledge Graph Embedding}
\acrodef{KGR}{Knowledge Graph Reasoning}
\acrodef{MDP}{Markov Decision Process}
\acrodef{GNN}{Graph Neural Network}

\copyrightyear{2025} 
\acmYear{2025} 
\setcopyright{rightsretained}
\acmConference[SIGIR '25]{Proceedings of the 48th International ACM SIGIR Conference on Research and Development in Information Retrieval}{July 13-18, 2025}{Padua, Italy}
\acmBooktitle{Proceedings of the 48th International ACM SIGIR Conference on Research and Development in Information Retrieval (SIGIR '25), July 13-18, 2025, Padua, Italy}
\acmISBN{979-8-4007-1592-1/25/07}
\acmDOI{10.1145/3726302.3730084}

\ccsdesc[500]{Computing methodologies~Discourse, dialogue and pragmatics}
\ccsdesc[300]{Human-centered computing~User centered design}

\keywords{User-tailored Dialogue Policy Planning, LLM-based Dialogue Agents}

\author{Tao He}\authornote{Work was done during an internship at SMU.}
\affiliation{%
  \institution{Harbin Institute of Technology}
  \city{Harbin}
  \state{Heilongjiang}
  \country{China}
}
\email{the@ir.hit.edu.cn}

\author{Lizi Liao}
\affiliation{%
  \institution{Singapore Management University}
  \city{Singapore}
  \country{Singapore}
}
\email{lzliao@smu.edu.sg}

\author{Ming Liu}\authornote{Corresponding Author: Ming Liu.}
\affiliation{%
  \institution{Harbin Institute of Technology}
  \city{Harbin}
  \state{Heilongjiang}
  \country{China}}
\email{mliu@ir.hit.edu.cn}

\author{Bing Qin}
\affiliation{%
  \institution{Harbin Institute of Technology}
  \city{Harbin}
  \state{Heilongjiang}
  \country{China}}
\email{qinb@ir.hit.edu.cn}

\renewcommand{\shortauthors}{Tao He et al.}

\begin{document}

\title{
Simulating Before Planning: Constructing Intrinsic User World Model for User-Tailored Dialogue Policy Planning
}

\begin{abstract}
Recent advancements in dialogue policy planning have focused on optimizing system agent policies to achieve predefined goals, emphasizing strategy design, trajectory acquisition, and training efficiency. However, these approaches often overlook the critical role of user characteristics, which are essential in real-world scenarios like conversational search and recommendation, where interactions must adapt to individual user traits such as personality, preferences, and goals. 
To address this gap, we conduct a comprehensive study using task-specific user personas to evaluate dialogue policy planning under diverse user behaviors. Our analysis, based on these user profiles, reveals significant shortcomings in existing approaches, underscoring the necessity for user-tailored dialogue policies.
Building on these insights, we propose the \textbf{U}ser-Tailored \textbf{D}ialogue Policy \textbf{P}lanning (UDP) framework, which integrates an Intrinsic User World Model to capture user traits and feedback. UDP operates in three stages: (1) User Persona Portraying, employing a diffusion model to dynamically infer user profiles; (2) User Feedback Anticipating, using a Brownian Bridge-inspired mechanism to predict user reactions; and (3) User-Tailored Policy Planning, synthesizing these elements to optimize response strategies. 
To enhance robustness, we introduce an active learning approach that prioritizes challenging user personas during training. Extensive experiments across benchmarks, including both collaborative and non-collaborative settings, demonstrate UDP’s effectiveness in learning user-specific dialogue strategies. Results confirm the framework’s utility, highlighting its robustness, adaptability, and potential to advance user-centric dialogue systems.
\end{abstract}
\maketitle

\acresetall
\section{Introduction}\label{sec:intro}
Tailoring interactions to individual user characteristics is essential for success in dialogue-based systems, particularly in applications like conversational search~\cite{tavakoli2022analyzing} and recommendation~\cite{christakopoulou2018q}. Users differ in their preferences, personalities, and decision-making styles, which shape their behaviors and expectations during interactions~\cite{Zhang2024StrengthLI, Scott1995DecisionMakingST}. 
To ensure effective dialogues, agents must perform strategic planning tailored to diverse users, dynamically adapting their strategies to align with individual traits and goals~\cite{Guo2024PCQPRPC, Dao2024ExperienceAS}.

Existing dialogue policy planning methods typically focus on optimizing interactions by improving system agent strategies through prompt engineering or externally trained planners. 
Prompt engineering aims to design detailed prompts to guide Large Language Models (LLMs) in decision-making~\cite{Yu2023PromptBasedMT, Deng2023PromptingAE}. 
Externally trained planners, on the other hand, use smaller, task-specific models to influence LLM behavior~\cite{Deng2023PlugandPlayPP, He2024PlanningLH}. 
Despite these advancements, these studies predominantly focus on enhancing the dialogue agent’s ability over the uniform user agent for different tasks, overlooking user-specific behavioral variability~\cite{Zhang2024StrengthLI}. 
Specifically, these methods rely on a fixed and uniform user prompt to simulate users~\cite{Deng2023PlugandPlayPP, He2024PlanningLH}, resulting in a more neutral role-playing style that aligns closely with the LLM's inherent ``default'' personality~\cite{Zhang2024StrengthLI}. However, handling different users with distinct backgrounds or preferences using different tactics is an essential skill for the experienced~\cite{He2024SimulationFreeHL}.


To investigate the limitations of current dialogue agents in handling diverse user traits, we conducted a comprehensive study utilizing task-specific user personas. These personas, designed for the P4G persuasion task~\cite{Wang2019PersuasionFG} and the ESConv emotional support task~\cite{Liu2021TowardsES}, were used to generate realistic user profiles that instruct LLMs to role-play diverse user behaviors during interactions~\cite{Lu2024LargeLM, Chen2024FromPT}. This approach enables a systematic assessment of dialogue systems' ability to adapt strategies across different user types (\textit{cf.} Section \ref{study} for details). Verification experiments revealed significant performance gaps, with an average difference of 19.2\% in P4G and 34.9\% in ESConv at SR metric for different baselines, highlighting two key challenges: (1) \textit{\textbf{making dialogue agents aware of diverse user characteristics}} and (2) \textit{\textbf{devising tailored strategies for individual users}}. 
These findings emphasize the importance of user diversity in dialogue policy planning and motivate the development of more advanced frameworks capable of delivering user-specific strategies effectively.

\begin{figure}[!t]
    \centering
    \includegraphics[width=\linewidth]{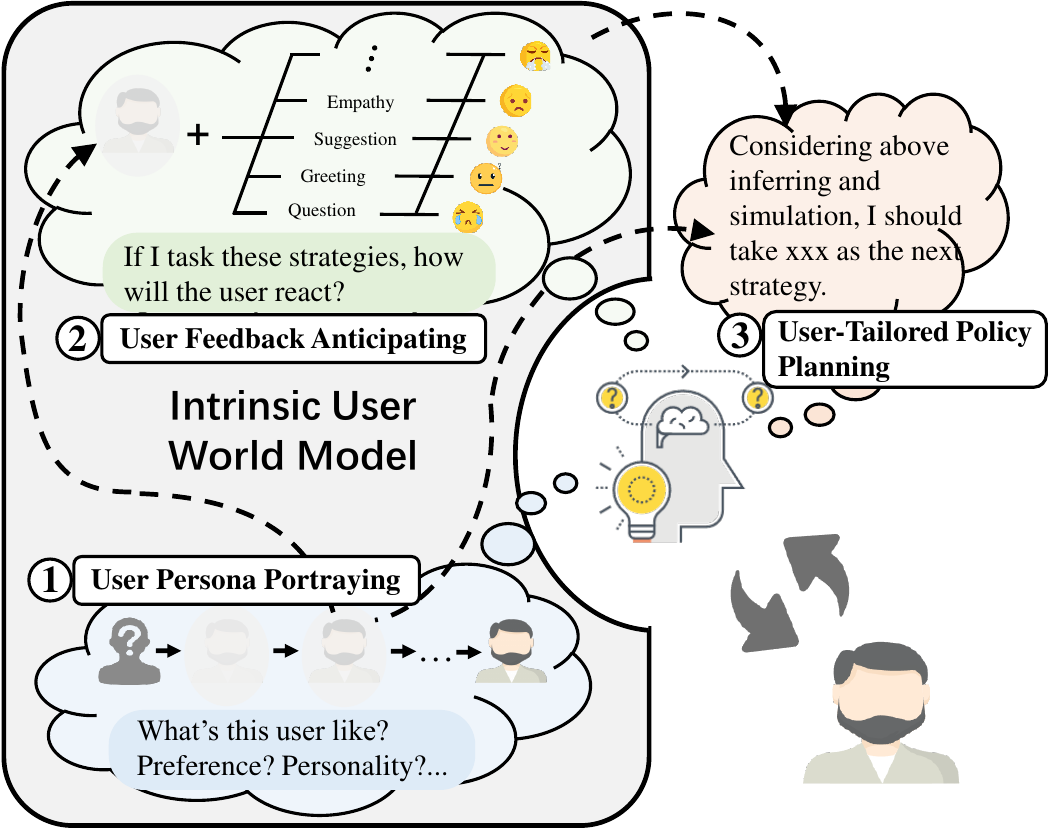}
    \caption{Framework overview. The framework UDP consists of three stages: User Persona Portraying, User Feedback Anticipating, and User-Tailored Policy Planning.}
    \vspace{-0.4cm}
    \label{fig:demonstration}
\end{figure}

To address the challenges, we propose \textbf{U}ser-Tailored \textbf{D}ialogue Policy \textbf{P}lanning (\textbf{UDP}), a novel framework that explicitly models user characteristics to enhance dialogue strategy planning. UDP operates in three stages: User Persona Portraying, User Feedback Anticipating, and User-Aware Policy Planning. In the first stage, the dialogue agent infers the user’s persona dynamically during interactions, modeled as a denoising task~\cite{Song2020DenoisingDI} using a Diffusion Model-based User Profiler. In the second stage, the agent anticipates user feedback to different strategies through a Brownian Bridge-inspired Feedback Anticipator~\cite{Revuz1990ContinuousMA}, enabling better alignment of actions with user expectations. The first two stages together form an intrinsic and simplified User World Model~\cite{Ha2018WorldM, LeCun2022APT}, allowing the agent to estimate user traits and potential future states~\cite{Ding2024UnderstandingWO}. 
In the final stage, the agent integrates dialogue history, inferred personas, and anticipated feedback to determine the optimal strategy. Moreover, we introduce an active online learning approach to refine UDP’s planning capabilities by initially focusing on all user personas and gradually emphasizing more challenging ones. Experiments demonstrate that UDP achieves consistent and significant performance improvements across diverse user personas, showcasing its robustness, adaptability, and generalizability.

To sum up, our main contributions are as follows:
\begin{itemize}
    \item We propose a novel evaluation protocol with task-specific user personas to systematically assess dialogue systems’ ability to adapt strategies for diverse user traits.
    \item We introduce User-Tailored Dialogue Policy Planning (UDP), a three-stage framework with an intrinsic User World Model to model user behaviors and optimize dialogue strategies.
    \item Comprehensive experiments on P4G and ESConv tasks demonstrate UDP’s significant performance improvements, validating its robustness and generalizability.
\end{itemize}
\section{Related Work}

\textbf{Policy Planning for LLM-based Dialogue Agents.}
Dialogue policy planning aims to determine the appropriate strategy before each response. 
In the LLM era, despite their impressive performance in dialogue tasks~\cite{Bang2023AMM, Zhao2023IsCE, Liang2024ActivelyLF}, LLMs still struggle with strategic planning in complex scenarios~\cite{Liu2021TowardsES}. Ask-an-Expert~\cite{Zhang2023AskAE} and ProCoT~\cite{Deng2023PromptingAE} use LLMs directly as the policy model via In-Context Learning to select policies, while GDP-Zero~\cite{Yu2023PromptBasedMT} further uses Monto Carlo Tree Search (MCTS) to improve planning. However, frozen LLMs lack domain-specific planning skills, and fine-tuning is costly. To address this, PPDPP trains a small planner to guide LLM responses using Reinfocement Learning algorithm~\cite{Sutton1999PolicyGM}. 
DPDP~\cite{He2024PlanningLH} enhances policy exploration efficiency using MCTS to guide planning. 
SDPP~\cite{He2024SimulationFreeHL} jointly learns policy planning and LLM policy-following behavior for better performance.
The aforementioned LLM-based approaches primarily focus on strategy modeling while neglecting the crucial user modeling.
In fact, user characteristics, such as personality traits, income, and other attributes, play the key role in shaping user responses and decisions during a conversation~\cite{Peng2018DeepDI, Zhang2019BudgetedPL}. Therefore, estimating user types and adopting tailored dialogue strategies accordingly is a basic skill owned by experienced persons.
The most relevant approach is TRIP~\cite{Zhang2024StrengthLI}, which prompts LLMs to predict user's potential behaviors.
However, we found that LLMs have limited capability in perceiving user traits via analysis experiments.
Instead, we develop a learnable user world model in the policy model to estimate user personas and predict reactions, based on Diffusion Models~\cite{Ho2020DenoisingDP} and Brownian Bridge Process~\cite{Revuz1990ContinuousMA}.

\textbf{User Simulation for Effective Dialogue Planning.}
Dialogue is an interactive activity where the user also plays an indispensable role. The user's response, as part of the new dialogue state, influences the next-turn strategy prediction. 
Therefore, the quality of user simulation greatly determines the learning quality of the dialogue agent. User simulation has been a consistent focus alongside the dialogue research.
Before LLM era, user simulation was approached through Agenda-based~\cite{Li2014TemporalSL} and Data-Driven methods~\cite{Kwan2022ASO}: the former used manually defined rules to constrain user feedback~\cite{Ultes2017PyDialAM, Gao2018NeuralAT, Zhang2020EvaluatingCR}, requiring extensive domain knowledge and lacking flexibility~\cite{Sekulic2024ReliableLU}; the latter learned user agents from data~\cite{Asri2016ASM, Lin2021DomainindependentUS, Lin2022GenTUSSU}, but demanded extensive annotated data. 
LLMs have simplified the user simulation subtask~\cite{Lin2023EmoUSSU, Chen2024TheOO}. 
Research has found that designing specific prompts enables LLMs to conduct vivid role playing~\cite{Li2023CAMELCA, Wang2023RoleLLMBE, Tu2024CharacterEvalAC}. 
Based on this, previous works explore a uniform prompt to guide user role-play across different conversation tasks~\cite{Deng2023PlugandPlayPP, He2024PlanningLH}. However, these methods overlook the impact of user traits and behavior diversity on policy planning.
To this end, TRIP~\cite{Zhang2024StrengthLI} designs 20 user persona prompts based on the BIG-5 Personality~\cite{Goldberg1992THEDO} and 4 Decision Types~\cite{Scott1995DecisionMakingST}. 
However, it lacks task-specific role simulation, and the user persona can be far from the task. In contrast, we design task-specific user attributes and prompts. We also design a novel framework to sense user characters and cater for better policy planning.
\section{User-Specific Planning Evaluation}
\label{study}
To systematically investigate the ability of dialogue agents to plan strategies tailored to diverse user characteristics, we conducted a comprehensive evaluation using our proposed protocol grounded in task-specific user personas. This evaluation aims to analyze how well existing dialogue agents adapt their strategies to user diversity and identify key challenges that arise when handling varied user traits in goal-oriented dialogue scenarios. 
Our study focuses on two representative tasks: P4G for persuasion~\cite{Wang2019PersuasionFG} and ESConv for emotional support~\cite{Liu2021TowardsES}, which represent non-cooperative and cooperative dialogue settings, respectively. The evaluation design consists of three key steps: (1) designing tailored user personas for each task, (2) generating detailed user profiles based on these personas, and (3) conducting evaluation experiments to validate the profile quality in simulating diverse user behaviors and the corresponding effects in affacting dialogue agents' task performance.

\subsection{User Persona Designing}
Different from previous work~\cite{Zhang2024StrengthLI}, which designs the same and task-agnostic user profiles for all tasks, we generate distinct user personas tailored to specific tasks for more realistic simulations.

The P4G task refers to persuading others to make a donation.
Considering the factors that impact decision results, including objective factors, subjective preferences, and personal decision-making habits, we designed user personas based on three key dimensions: \textbf{income}, \textbf{personality}, and \textbf{decision type}.
Each dimension is divided into two distinct properties: [Well-off, Financially-tight], [Compassionate, Rejective], and [Rational, Emotional]. 
Besides, we also define ``hard'' and ``easy'' donation acceptance difficulty for each user persona.
This adjustment results in a total of $2^3 \times 2 = 16$ user personas for P4G. 
For ESConv, we also designed the user persona from the user-centric perspective, considering three key aspects: {communication ability} (affecting the clarity of user self-expression), {emotional state} (influencing whether the user requires more suggestions or reassurance), and {willingness to accept suggestions} (determining the likelihood of adopting the provided suggestions).
Specifically, we designed user personas based on three key dimensions: \textbf{extroversion}, 
\textbf{openness}, 
and \textbf{neuroticism}.
Each dimension includes two distinct attributes: [Extravert, Introvert], [Adventurous, Conservative], and [Rational, Neurotic]. 
Combining these dimensions yields $2^3 = 8$ distinct user personas.

\subsection{User Profile Generation}
Similar to TRIP~\cite{Zhang2024StrengthLI}, we rely on the instruction-following ability of LLMs to generate specific user profiles based on the designed personas. We employ GPT-3.5-turbo and provide detailed characteristics that guide it in role-playing diverse user behaviors during interactions.
For P4G, we generate 400 unique profiles for training (25 per persona) and 80 profiles each for validation and testing (5 per persona). Since P4G lacks predefined dialogue contexts, these profiles ensure variability across dialogue cases. For ESConv, we generate 40 profiles for training (5 per persona) and 8 profiles each for validation and testing.

\subsection{Verify Profile Quality \& Performance Gap}
To validate the quality of the generated profiles, we conducted verification experiments using these profiles on the dialogue planning task. 
Our validation experiment consists of two parts: first, we aim to show that these profiles can effectively guide the LLM in role-playing through human evaluation; second, by analyzing dialogue success rate differences across various profiles on the policy planning task using automatic metrics, we aim to demonstrate the impact of user diversity on the planning performance.

\subsubsection{Self-play Dialogue Interaction.}
To conduct the following evaluations, we first need to solve how to simulate dialogues using these user profiles.
In line with prior work~\cite{Deng2023PlugandPlayPP}, we employ three LLMs to serve as the system, user, and critic. 
We use GPT-4o-mini to play the roles of the system and user during multi-turn dialogue simulations, respectively. 
Different from prior works, we equipped the user agent with distinct traits, personalities, and decision styles guided by various user profiles generated before. 
After each dialogue turn, we prompt GPT-3.5-turbo to act as a critic, assessing whether the objective has been met and providing a score:
\begin{itemize}
\item For P4G, we prompt the LLM with a multiple-choice question: ``Has the Persuadee agreed to donate to the charity?'' The options are ``refused'', ``neutral'', ``positive'', and ``agree'', mapped to values of -1.0, -0.5, 0.1, and 1.0, respectively.
\item For ESConv, we prompt the LLM with another question: ``What’s the Patient’s current emotional state?'' The options include ``worse'', ``same'', ``better'', ``accepted'' and ``solved'', also corresponding to values of -1.0, -0.5, 0.1, 1.0, and 1.0.
\end{itemize}
For the critic, we set the sampling temperature $\tau=1.1$ and generate 10 options. These options are mapped to a scalar value $r(s_t)$, representing the current dialogue completion state. Following previous work~\cite{He2024SimulationFreeHL}, we view the dialogue objective as being achieved if $r(s_t)$ exceeds a threshold of 0.6. 
The simulated dialogue continues for multiple turns until reaching the maximum dialogue turn (10) or the dialogue goal is achieved.

\subsubsection{Human Evaluation on Profile Quality}
Role-playing via profile guidance for LLMs has been verified in prior studies~\cite{Ye2024SweetieChatAS, wu2025raiden, Chen2024FromPT}. 
To evaluate whether our designed user profiles can effectively guide LLMs in role-playing, ensuring LLMs' behaviors align with the profiles, we conducted a human evaluation.
We first generated 80 dialogues (10 per persona) on P4G and 40 (5 per persona) for ESConv through the self-play interaction, with each dialogue containing a predefined user persona and profile. Three annotators were required to evaluate whether the user’s utterances kept consistent with the traits described in the user profile. For ESConv, annotators evaluated the attributes of extroversion, openness, and neuroticism; for P4G, they evaluated income, personality, and decision type. We calculated the mean of three evaluated consistency results.

\noindent\textbf{Evaluation Results.}
In P4G, the consistency degrees of generated dialogues with the user profiles achieve 95.8\%, 100\%, and 96.25\% for the income, preference, and decision type attributes, respectively. In ESConv, the consistency degrees reach 85\%, 90.83\%, and 98.33\% for the extroversion, openness, and neuroticism. 
The Fleiss Kappa~\cite{Moons2023MeasuringAA} values for 3 annotators across different tasks and personas all exceed 0.4.
These results show ChatGPT's capability to accurately adhere to the specified user profiles during role-playing, thereby validating the effectiveness of the designed profiles and paving the way for the following studies.

\subsubsection{Automatic Evaluation on Task Performance}\label{sec:automatic_evaluation}
We further conduct automatic evaluation of policy planning tasks on existing baselines.
We selected two representative methods—ProCoT~\cite{Deng2023PromptingAE}, a prompt-based planner, and PPDPP~\cite{Deng2023PlugandPlayPP}, a trained small model-based planner—to analyze performance variance across various user profiles. 
During self-play, these planners first predict the appropriate dialogue strategies, which then guide the system in generating specific responses.
Following previous work~\cite{He2024SimulationFreeHL}, we consider two metrics for this task: Success Rate (\textbf{SR}) and Soft Success Rate (\textbf{SSR}). 
The SR metric measures the dialogue success rate by calculating the percentage of successful dialogues within all test cases. 
And the SSR metric computes the mean of final-turn rewards in the simulated dialogues for all test cases.

\noindent\textbf{Evaluation Results.}
As presented in Table~\ref{tab:preliminary_exp}, dialogue performance varies significantly across user personas, with a maximal average SR difference of 19.2\% in P4G and 34.9\% in ESConv, underscoring the impact of different user traits on interaction outcomes.

Key findings from the results include: (1) Lack of User Awareness: Existing dialogue agents struggle to accurately infer and adapt to diverse user characteristics, leading to generic interactions and reduced task success. (2) Inadequate Strategy Adaptation: Dialogue agents fail to personalize strategies effectively based on user personas, resulting in suboptimal user experiences.
These insights validate the quality of our designed user profiles in effectively simulating distinct behavioral patterns. Moreover, they highlight critical challenges in user-specific policy planning and emphasize the need for advanced frameworks that can dynamically estimate user traits and optimize dialogue strategies.

\begin{table}[!htp]
    \caption{The performance comparison across different user personas for ProCoT and PPDPP on P4G and ESConv.}
    \vspace{-0.35cm}
    \centering
    \subtable[P4G on persuasion dialogues.]{
        \resizebox{0.92\linewidth}{!}{
        \begin{tabular}{cccccc}
        \toprule
        \multicolumn{2}{c}{\multirow{2}{*}{\textbf{Persona}}} & \multicolumn{2}{c}{\textbf{ProCoT}} & \multicolumn{2}{c}{\textbf{PPDPP}} \\
        \cmidrule{3-6}
        & & SSR$\uparrow$ & SR$\uparrow$ & SSR$\uparrow$ & SR$\uparrow$ \\
        \midrule
        \multirow{2}{*}{\textbf{Income Level}} & \multicolumn{1}{l}{Stable} & 0.490 & 0.625 & 0.441 & 0.600 \\
        & \multicolumn{1}{l}{Poor} & 0.303 & 0.525 & 0.323 & 0.575 \\
        & \multicolumn{1}{l}{$|\Delta|$} & 0.187 & 0.100 & 0.118 & 0.025 \\
        \midrule
        \multirow{2}{*}{\textbf{Preference}} & \multicolumn{1}{l}{Compassionate} & 0.673 & 0.775 & 0.685 & 0.825 \\
        & \multicolumn{1}{l}{Rejective} & 0.120 & 0.375 & 0.079 & 0.350 \\
        & \multicolumn{1}{l}{$|\Delta|$} & 0.553 & 0.400 & 0.606 & 0.475 \\
        \midrule
        \multirow{2}{*}{\textbf{Decision Type}} & \multicolumn{1}{l}{Rational} & 0.362 & 0.550 & 0.348 & 0.550 \\
        & \multicolumn{1}{l}{Emotional} & 0.431 & 0.600 & 0.416 & 0.625 \\
        & \multicolumn{1}{l}{$|\Delta|$} & 0.069 & 0.050 & 0.068 & 0.075 \\
        \midrule
        \multicolumn{2}{c}{\textbf{Average $|\Delta|$}} & \textbf{0.270} & \textbf{0.183} & \textbf{0.264} & \textbf{0.192} \\
        \bottomrule
        \end{tabular}
        }
    }
    \subtable[ESConv on emotion support dialogues.]{
        \resizebox{0.92\linewidth}{!}{
        \begin{tabular}{cccccc}
        \toprule
        \multicolumn{2}{c}{\multirow{2}{*}{\textbf{Persona}}} & \multicolumn{2}{c}{\textbf{ProCoT}} & \multicolumn{2}{c}{\textbf{PPDPP}} \\
        \cmidrule{3-6}
        & & SSR$\uparrow$ & SR$\uparrow$ & SSR$\uparrow$ & SR$\uparrow$ \\
        \midrule
        \multirow{2}{*}{\textbf{Openness}} & \multicolumn{1}{l}{Open} & 0.554 & 0.621 & 0.851 & 0.924 \\
        & \multicolumn{1}{l}{Conservative} & 0.228 & 0.359 & 0.550 & 0.563 \\
        & \multicolumn{1}{l}{$|\Delta|$} & 0.326 & 0.262 & 0.301 & 0.361 \\
        \midrule
        \multirow{2}{*}{\textbf{Extravertion}} & \multicolumn{1}{l}{Extravert} & 0.655 & 0.727 & 0.805 & 0.879 \\
        & \multicolumn{1}{l}{Introvert} & 0.123 & 0.250 & 0.598 & 0.609 \\
        & \multicolumn{1}{l}{$|\Delta|$} & 0.532 & 0.477 & 0.207 & 0.270 \\
        \midrule
        \multirow{2}{*}{\textbf{Neuroticism}} & \multicolumn{1}{l}{Rational} & 0.599 & 0.646 & 0.827 & 0.892 \\
        & \multicolumn{1}{l}{Neurotic} & 0.188 & 0.338 & 0.579 & 0.600 \\
        & \multicolumn{1}{l}{$|\Delta|$} & 0.411 & 0.308 & 0.248 & 0.292 \\
        \midrule
        \multicolumn{2}{c}{\textbf{Average $|\Delta|$}} & \textbf{0.423} & \textbf{0.349} & \textbf{0.252} & \textbf{0.308} \\
        \bottomrule
        \end{tabular}
        }
    }
    \vspace{-0.3cm}
    \label{tab:preliminary_exp}
\end{table}
\section{Methodology}
\subsection{Preliminaries}
\subsubsection{Problem Formulation}
Following existing work~\cite{Deng2023PlugandPlayPP, Zhang2024StrengthLI}, we formulate the dialogue policy planning task as a Markov Decision Process (MDP): $(\mathcal{S}, \mathcal{A}, r, \mathcal{T})$, where $\mathcal{S}$ is the dialogue state (history) space; $\mathcal{A}$ is the dialogue action (strategy) space, i.e., the set of candidate dialogue strategies pre-defined by domain experts; $r:\mathcal{S}\times\mathcal{A}\times\mathcal{S}\rightarrow \mathbb{R}$ is the reward function;
and $\mathcal{T}: \mathcal{S}\times\mathcal{A}\rightarrow\mathcal{S}$ is the transition function. 
At each turn $t$, according to the current state $s_t\in\mathbb{S}$, i.e., the dialogue history, 
the dialogue planner selects a strategy $a_{t+1}\in \mathcal{A}$. Then, guided by it, the system player generates the utterance $u^{sys}_{t+1}$. In return, the user player responds with $u^{usr}_{t+1}$. After the $t+1$-th turn is finished, the dialogue state transitions to $s_{t+1}=\mathcal{T}(s_t,a_t)=\{u^{sys}_1, u^{usr}_1,..., u^{sys}_{t+1}, u^{usr}_{t+1}\}$, and a reward $r_{t+1}$ is accessed by the critic player. 
This process repeats until the dialogue goal is achieved or the maximum number of turns $T$ is reached. 
Our target is to learn a policy $\pi_\theta$ maximizing the expected cumulative rewards over observed dialogue episodes as:
\begin{equation}
    \pi^*=\arg \max_{\pi_\theta}\bigg[\sum_{t=0}^Tr(s_t,a_t)\bigg].
    \label{eq:total_target}
\end{equation}

\subsubsection{Diffusion Probabilistic Models}\label{sec: diffusion_model_introduction}
Diffusion probabilistic models~\cite{Ho2020DenoisingDP, Song2020DenoisingDI} are generative models that first slowly corrupt the structure in data $\mathbf{x}_0\sim q(\mathbf{x}_0)$ by adding noise through a forward diffusion process $q(\mathbf{x}_i|\mathbf{x}_{i-1})$ defined as:
\begin{equation}
    q(\mathbf{x}_{i-1}|\mathbf{x}_i)=\mathcal{N}(\mathbf{x}_i;\sqrt{1-\beta_i}\mathbf{x}_{i-1},\beta_i\mathbf{I}),
\end{equation}
where $\mathcal{N}$ is the Gaussian distribution, $i\in \{0,1,...,N\}$ is the diffusion timestep, which is distinguished from the dialogue turn step $t$. $\{\beta_i\}_{i=1}^N$ is the predefined hyperparameters to control the noising degrees per step.
Then, a trainable de-noising procedure $p_\phi(\mathbf{x}_{i-1}|\mathbf{x}_i)$ is constructed as:
\begin{equation}
    p_\phi(\mathbf{x}_{i-1}|\mathbf{x}_i)=\mathcal{N}(\mathbf{x}_{i-1};\mu_\phi(\mathbf{x}_i,i),\Sigma_i),
\end{equation}
where $\mu_\phi(\mathbf{x}_i, i)$ is the forward process posterior mean, implemented using a function of a noise prediction neural network $\epsilon_\phi(\mathbf{x}_i, i)$ with a learnable parameter $\phi$.
In this work, we consider conditional diffusion probabilistic models~\cite{Dhariwal2021DiffusionMB, Xiang2024DiffusionDialogAD}, which represent a distribution $p(\mathbf{x}_0|c)$ over a dataset of samples $\mathbf{x}_0$ and corresponding context $c$. 
It aligns well with dialogue systems as they progressively refine noisy inputs, similar to how dialogue agents gradually infer user characteristics over interactions. Based on this, we adopt a diffusion-based model to design the User Portrayer, enabling structured and incremental user modeling for personalized dialogue planning.

\subsubsection{Brownian Bridge Stochastic Process}
Brownian Bridge Process~\cite{Revuz1990ContinuousMA} is a continuous-time Gaussian stochastic process $B(t)$ whose probability distribution of each time step follows a Gaussian distribution and is conditioned by start state $z_0$ at $t=0$ and end state $z_T$ at $t=T$:
\begin{equation}
    B(t)=\mathcal{N}((1-\frac{t}{T}){z}_0+\frac{t}{T}{z}_T, \frac{t(T-t)}{T})\quad t\in [0,T].
    \label{eq: brownian_bridge}
\end{equation}
An empirical comprehension is that ${z}_t\sim B(t)$ is approximately the noisy linear interpolation of $z_0$ and $z_T$ modulated by time variable. The uncertainty gradually decreases to the lowest at the start side or end side of the bridge and increases to the highest at the bridge center point.
A useful property of the Brownian Bridge Process is that if there exists a point $0<\tau<T$ such that $z_0 < z_\tau < z_T$, then the equation in Eq.\ref{eq: brownian_bridge} is also satisfied within $[\tau, T]$.

The Brownian Bridge Process naturally models state uncertainty and is easy to implement. It has been shown valid in dialogue scenarios~\cite{Wang2023DialoguePV}. While COLOR~\cite{Wang2023DialoguePV} uses the Brownian Bridge Process to model the policy planning process of the dialogue agent, we instead employ it to learn the user's feedback state.

\subsection{Our \textit{UDP} Framework}
To perceive user personas and anticipate user behaviors during the dialogue and tailor policies for different user personas, we propose the \textbf{UDP} framework, inspired by human-like thinking mode when interacting with users exhibiting diverse traits and behaviors.
This framework comprises three stages: \textbf{User Persona Portraying}, \textbf{User Feedback Anticipating}, and \textbf{User-Aware Policy Planning}. 
As illustrated in Figure~\ref{fig:demonstration}, the first two stages collaboratively construct an intrinsic and simplified world model that enables the dialogue agent to internally simulate the potential consequences of adopting different strategies. 
This process starts with inferring ``what the user might be like'' as foundational information and progresses to simulating ``how the user might react if I respond in this way.'' 
Finally, the agent formulates the next-step strategy based on the inferred user persona and the simulated user responses.

\subsubsection{\textbf{Stage 1: User Persona Portraying}}

\begin{figure}[!t]
    \centering
    \includegraphics[width=0.9\linewidth]{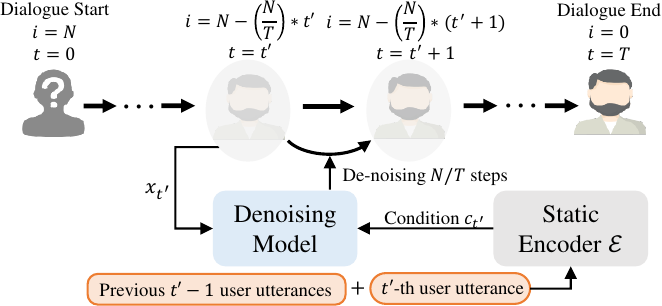}
    \caption{The portraying (denoising) process of the Diffusion Model-based User Persona Portraying stage.}
    \vspace{-0.4cm}
    \label{fig:diffusion_model}
\end{figure}

We observe that during a dialogue, the agent's understanding of user attributes evolves gradually from an initial state of poor awareness to increasing clarity. This process aligns well with the denoising process in the diffusion model, which transitions from pure noise to generating realistic samples. Based on this observation, we propose a condition-based diffusion model for user profiling, as illustrated in Figure~\ref{fig:diffusion_model}.

At the $i = N$ step, the diffusion process is initialized with complete noise $\mathbf{x}^N$, and each denoising step progressively reduces the noise. After $N$ steps, the process reaches a noise-free state $\mathbf{x}^0$. In the task of estimating user traits, at $t = 0$, the dialogue agent has no prior knowledge of the user. As the dialogue going, each user response incrementally informs the agent’s understanding of the user. After $T$ dialogue turns, the agent’s understanding of the user is maximized. We model this process of trait estimation as a denoising procedure, where the agent’s initial lack of knowledge corresponds to the fully noisy state, and the final understanding at the end of the dialogue corresponds to the denoised state. Since $N \gg T$ (with $N = 1000$, $T = 10$ in this work), we perform $N/T$ denoising steps after each dialogue turn.

Specifically, the denoising process starts with $\mathbf{x}_0^N \sim \mathcal{N}(0,1)$, representing the state of complete unawareness at $t = 0$. After the $t$-th dialogue turn, the agent has received prior $t$ user responses $u_{1:t} = [u_1^{usr}, \dots, u_t^{usr}]$ and encode it into $\mathbf{u}_{1:t}$ using a frozen RoBERTa-base model $\mathcal{E}$. Then the agent performs $N/T$ denoising steps based on $\mathbf{u}_{1:t}$. Each denoising step is modeled as $p(\mathbf{x}_t^{i-1} | \mathbf{x}_t^i, c = \mathbf{u}_{1:t})$, where $i \in [N(1 - t/T), N(1 - (t+1)/T))$. Upon receiving the final $T$-th user response, the denoising process proceeds from $\mathbf{x}_T^{N/T}$ to the final noise-free state $\mathbf{x}_T^0$ after $N/T$ denoising steps.

It is important to note that the state $x_t^i$ obtained after $i$ denoising steps is not directly used for estimating the user's traits for subsequent tasks. Instead, user attributes are estimated using the following formula: $\hat{\mathbf{x}}^0 = \frac{1}{\sqrt{\bar{\alpha}_i}} \mathbf{x}_t^i - \frac{\sqrt{1 - \bar{\alpha}_i}}{\sqrt{\bar{\alpha}_i}} \hat{\epsilon}_t$, where $\hat{\epsilon}_t$ is the noise estimation network $\epsilon_\phi(.)$ introduced in Section~\ref{sec: diffusion_model_introduction}.
By computing the similarity between $\hat{\mathbf{x}}^0$ and a set of pre-encoded user persona representations $\mathbf{P} = [\mathbf{p}_i]_{i=1}^M$ ($M$ is the number of user personas), we derive the user persona prediction distribution $D_t$. $\mathbf{p}_i$ is derived by encoding the user persona $p_i$ using the frozen encoder $\mathcal{E}$. 
During evaluation, we select the user persona with the highest probability from $D_t$ to help subsequent dialogue policy planning.

\subsubsection{\textbf{Stage 2: User Feedback Anticipating}}
A skilled negotiator considers the counterpart's potential future reactions before responding, avoiding invalid strategies. This motivation aligns theoretically with the core idea in model-based RL~\cite{Yi2018ModelbasedRL, Kaiser2019ModelBasedRL}, where the agent builds an internal model for the environment (i.e., the user in this task) to guide decision-making. 
To this end, our framework involves the user feedback anticipating module $Pt$ that can simulate various user reactions to different system strategies, providing crucial insights for policy planning. 
To balance efficiency and cost, we design $Pt$ based on the Brownian Bridge Process.

In our work, we define the initial state of the Brownian Bridge Process at $t=0$ (dialogue start) as $\mathbf{z}_0 = \overrightarrow{0}$, and the terminal state at $t=T$ (dialogue end) as the user’s persona, denoted by $\mathbf{z}_T = f_P(\mathbf{p}_i)$, where $f_P$ denotes a nonlinear map implemented by an MLP layer. 
Therefore, for $M$ user types, we establish $M$ distinct Brownian Bridge Processes. Our foundational assumption is that different user personas focus on distinct concerns, while users with the same persona tend to exhibit similar behavior trajectories. A similar assumption was proposed in COLOR~\cite{Wang2023DialoguePV}, where a Brownian Bridge Process is used to model the strategy trajectory of the system agent. 
Instead, our approach focuses on predicting user reactions.
During the pretraining phase of this module, we have access to the given user persona. While for inference, we use the user persona estimated in the User Persona Portraying stage.

\begin{figure}[!t]
    \centering
    \includegraphics[width=0.85\linewidth]{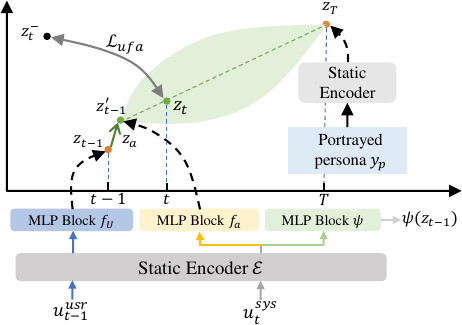}
    \vspace{-0.1cm}
    \caption{The architecture of the Brownian Bridge Process-based User Feedback Anticipating stage.}
    \vspace{-0.3cm}
    \label{fig:brownian_bridge}
\end{figure}

Leveraging the property of the Brownian Bridge Process that it still holds over the subset $[\tau,T]$ mentioned in Preliminaries, we can predict the user’s next-turn reaction. Specifically, before continuing the $t$-th turn dialogue, we have already observed the $t-1$-th user's response $u_{t-1}^{usr}$ and encodes it into $\mathbf{u}_{t-1}^{usr}$. We hypothesize the system agent takes the strategy $a_{t}$ in the next turn. 
Following the transition distribution form of the Brownian Bridge Process given in Eq.\ref{eq: brownian_bridge}, we derive the distribution of possible user responses to system utterance $u_{t}^{sys}$ at the $t$-th turn as follows:
\begin{equation}
    \begin{aligned}
p(\mathbf{z}_{t})\sim\mathcal{N}\big(\underbrace{\frac{(T-t)(\mathbf{z}_{t-1}+\mathbf{z}_a)}{T-t+1}+\frac{\mathbf{z}_T}{T-t+1}}_{\mu_t},
        \underbrace{\frac{4(T-t)\psi(\mathbf{z}_a)}{(T-t+1)^2}}_{\sigma_t^2}\big),
        \label{eq:bbp_equ}
    \end{aligned}
\end{equation}
where $\mathbf{z}_{t-1}=f_U(\mathbf{u}_{t-1}^{usr})$ represents the process state at time $t-1$. $\psi(.)$ is implemented using an MLP to calculate the variance of the distribution. 
Inspired by the previous method, we introduce $\mathbf{z}_a=f_a(\mathbf{a}_{t})$ ($\mathbf{a}_t$ is the feature vector of the action $a_t$) to perturb the density of the Brownian Bridge Process, accounting for differences in user responses based on different agent strategies. 
$f_U(.)$ and $f_a(.)$ mentioned above are also MLPs.

\subsubsection{\textbf{Stage 3: User-Aware Policy Planning}}
Ultimately, in the planning phase, the agent considers both the user’s profile and potential user reacts for each possible strategy. By integrating all available information, the agent determines the optimal strategy.

\begin{figure}[!t]
    \centering
    \subfigure[PPDPP \& TRIP]{
        \includegraphics[width=0.23\linewidth]{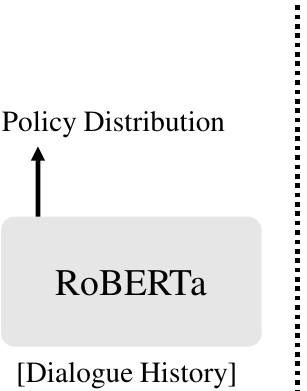}
    }
    \subfigure[UDP]{
        \includegraphics[width=0.72\linewidth]{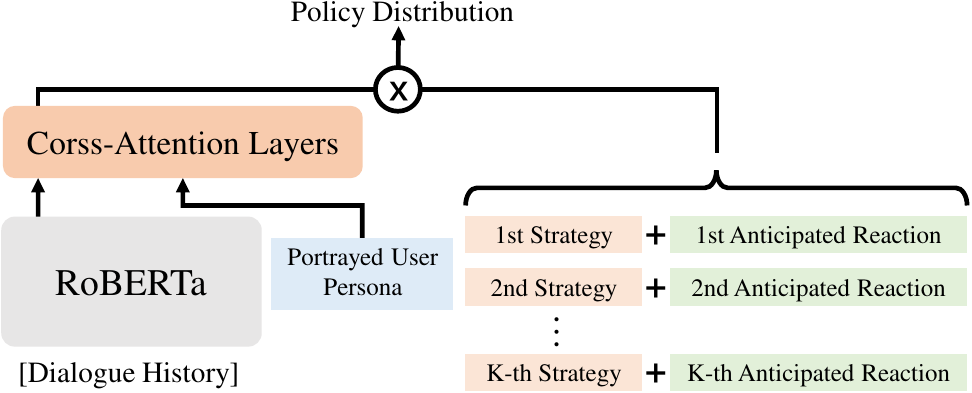}
    }
    \vspace{-0.1cm}
    \caption{The architecture comparison of the Policy Planning stage: (a) PPDPP \& TRIP; (b) UDP (Our).}
    \vspace{-0.3cm}
    \label{fig:planner}
\end{figure}

To achieve this, we propose a user-aware policy planner $Pa$, as illustrated in Figure~\ref{fig:planner}. First, the dialogue history $s_{t-1}$ over the prior $t$ turns is encoded to obtain $\mathbf{s}_{t-1}$ using the static encoder $\mathcal{E}$. Next, we feed $\mathbf{s}_{t-1}$ along with user persona features $\mathbf{p}$ into a multi-layer Transformer, yielding a fused representation of dialogue state and user personas $\mathbf{sp}_{t-1}\in \mathbb{R}^{1 \times d}$, where $d$ is the output dimension of $\mathcal{E}$. 
Each candidate strategy feature $\mathbf{a}_i$ is concatenated with corresponding predicted user reaction $\mathbf{z}_i$, forming an action feature $A_i = \text{MLP}([\mathbf{a}_i; \mathbf{z}_i]) \in \mathbb{R}^d$, thus defining the action space representation $\mathbf{A} = [A_i]_{i=1}^K \in \mathbb{R}^{K \times d}$, where $K$ is the size of the strategy set. Finally, we calculate the sampling probability for each action:
\begin{equation}
    \pi_\theta(a_t|s_{t-1})=Softmax(\mathbf{sp}_{i-1}\cdot \mathbf{A}^T),
\end{equation}
where $\theta$ refer to the parameters of the policy planner.

\subsection{Optimization}
To achieve faster convergence, we first pretrain these three stages before conducting online learning. To this end, we constructed a set of pretraining data. 
Specifically, we utilize the generated user profiles as prompts for user role play and adopt the self-play method to simulate dialogues, which is introduced in Section~\ref{sec:automatic_evaluation}. 
For the 400 user profiles of P4G, we created two dialogues per profile, resulting in 800 dialogues.
For the 40 user profiles of ESConv, we sampled 10 different user issues/situations from the training set for each profile, producing 400 dialogues. 
These dialogues were prompted using the predefined user profiles; thus, the constructed pretraining dataset inherently contains user persona labels.

\subsubsection{Pretraining User Persona Portrayer}
We pretrain the portrayer module using the constructed pretraining dataset, which includes user persona labels for each dialogue.
Unlike typical generative tasks where diffusion models are commonly applied, our user-profile portraying task is a classification problem. Therefore, instead of applying the loss function defined in DDPM~\cite{Ho2020DenoisingDP} to train this module, we leverage the predicted user persona distribution $D_t$ along with the true user persona label $y_p$ to train this module:
\begin{equation}
    \mathcal{L}_{upp}(\phi)=\text{CrossEntropyLoss}(D_t,y_p).
\end{equation}

\subsubsection{Pretraining User Feedback Anticipator}
We also pretrain this module based on the pretraining dataset above.
For a training example $S=(u_{t-1}^{usr},u_t^{sys},u_t^{usr},y_p)$, in which $y_p$ is the persona label index, we first use Eq.\ref{eq:bbp_equ}  
to predict the process state $\hat{\mathbf{z}}_{t}$ at time $t$ using the expectation $\mu_t$. We then compute the golden state at time $t$ as $\mathbf{z}_t^+=f_U(\mathbf{u}_{t}^{usr})$. 
The intuition behind the training process is to ensure that the predicted user feedback $\hat{\mathbf{z}}_t=µ_t$ is close to $\mathbf{z}_t^+$.
Following prior work~\cite{Wang2023DialoguePV}, we adopt a contrastive learning method to train this module. Formally, given a randomly sampled batch $B$, the optimization objective is defined as:
\begin{equation}
    \begin{aligned}
        \mathcal{L}_{ufa}(\xi)=-\log\frac{\exp(d(\hat{\mathbf{z}}_t;\mathbf{z}_t^+))}{\sum_{z\in B}\exp(d(\hat{\mathbf{z}}_t;\mathbf{z}))}\\
        d(\hat{\mathbf{z}}_t;\mathbf{z})=-\frac{1}{2\sigma_t^2}\|\hat{\mathbf{z}}_t-\mathbf{z}\|^2_2,
    \end{aligned}
\end{equation}
where $\sigma_t^2$ denotes the variance as defined in Eq.\ref{eq:bbp_equ}.

\subsubsection{Pretraining User-Aware Policy Planner}
After pretraining the world model, i.e., the first 2 modules, we freeze them and proceed to pretrain the 3rd stage. To pretrain the policy planner module, we first employ ChatGPT to annotate the system utterance $u_{t}^{sys}$ in each dialogue turn with a corresponding strategy $a_{t}$ from the given strategy set. 
For the two datasets, we sampled 50 strategy annotations and conducted a simple evaluation of the annotation accuracy. We find that the annotation quality is within an acceptable range.
Based on the constructed dataset $\{(s_t, a_t)\}$, we conduct supervised pretraining as follows:
\begin{equation}
    \mathcal{L}_{uapp}(\theta)=-\sum_{t=1}^T\log \pi_\theta(a_t|s_{t-1}).
\end{equation}

\subsubsection{Active Learning-based Optimization}
After pretraining, we conduct online RL to further enhance the agent’s planning capacity through exploration. 
We prompt three LLMs as the user, system, and critic to perform self-play conversations.
At each turn $t$, the planner selects a dialogue strategy according to $\pi_\theta(a_t|s_t)$, guiding the system LLM to generate a response $u^{sys}_{t}$. 
The simulated user then provides a reply $u^{usr}_{t}$, and the critic evaluates the dialogue state to assign a reward $r_t$ for the strategy $a_t$. 
This process continues until the dialogue goal is achieved or the maximum turn is reached. After the interaction, we train the agent using a vanilla policy gradient method following prior work. The formalization of this training objective is as follows:
\begin{equation}
    \mathcal{L}_{rl}(\theta)=-\sum_{t=1}^T\log\pi_\theta(a_t|s_{t-1})\cdot R_t,
\end{equation}
where $R_t = \sum_{t'=t}^T \gamma^{T-t'} r_{t'}$ represents the cumulative reward.
During online training, we focus exclusively on optimizing the Planner while keeping the Portrayer and Anticipator modules fixed.

Before conducting self-play interactions, we need to first choose a user profile to prompt the user LLM in role-playing.
Results in Table~\ref{tab:preliminary_exp} illustrate that the difficulty of achieving success varies across these user profiles. To enable the agent to trial more times with more challenging users, we propose an active learning-based online training method, allowing the agent to learn with more difficult users.
Specifically, for all user personas $\{{p}_i\}_{i=1}^M$, we assign a sampling weight $\{w_i\}_{i=1}^M$ to sample the user persona. The initial weights are set equally for all user personas. After each multi-turn dialogue simulation, we update the weights $\{w_i\}_{i=1}^M$ based on the employed user persona $p_j$ and the outcome of the conversation (whether successful or not), formulated as follows:
\begin{equation}
    w_i=
\begin{cases}
w_i+\mathbb{I}(i==j),  & \text{ if the dialogue is successful}; \\
w_i-\mathbb{I}(i==j), & \text{    otherwise}. \\
\end{cases}
\end{equation}
\section{Experiments}
\subsection{Experimental Setup}
We follow a similar experimental setup as the previous Validation Experiments in Section~\ref{sec:automatic_evaluation}, except for the following differences:
\subsubsection{Baselines}
We consider comparing two main categories of methods: (1) \textbf{Prompt-based} methods, including \textbf{Standard Prompt} (where the LLM directly acts as the system without considering dialogue strategy), \textbf{Proactive}~\cite{Deng2023PromptingAE} (where the LLM first selects a policy and then generates a response accordingly), \textbf{ProCoT}~\cite{Deng2023PromptingAE} (where the LLM uses Chain-of-Thought (CoT) reasoning to generate the next policy), \textbf{ICL-AIF}~\cite{fu2023improving} (where another LLM acts as a coach, analyzing the dialogue state and guiding the system on how to respond following), and \textbf{GDP-Zero}~\cite{Yu2023PromptBasedMT} (where MCTS is conducted to help search the dialogue policy); and (2) \textbf{Planner-based} methods, including \textbf{PPDPP}~\cite{Deng2023PlugandPlayPP} (where RoBERTa is trained as a planner to select policies) and \textbf{TRIP}~\cite{Zhang2024StrengthLI} (which uses the LLM to infer the user’s mental states and next actions, providing insights for the planner to select policies). It is important to note that TRIP does not infer user persona types or consider user personas when anticipating the following user behaviors.

\subsubsection{Metrics}
Besides \textbf{SR} and \textbf{SSR} introduced in Section~\ref{sec:automatic_evaluation}, we also include the commonly used metric Average Turn (\textbf{AvgT}), which evaluates dialogue efficiency by calculating the average number of turns across simulated dialogues for all evaluation cases. As in the prior study~\cite{Zhang2024StrengthLI}, we set the maximum dialogue turns as 10. If the dialogue goal is achieved, the self-play will finish prematurely.

\begin{table}[t]
    \caption{Experimental results on P4G and ESConv. PT means Pretraining and RL means Reinforcement Learning. Results are averaged over five times inference (p < 0.05 under t-test)}
    \vspace{-0.2cm}
    \centering
    \renewcommand*{\arraystretch}{1.1}
    \resizebox{\linewidth}{!}{
    \begin{tabular}{lcccccc}
    \toprule
    \multirow{2}{*}{\textbf{Models}} & \multicolumn{3}{c}{\textbf{P4G}} & \multicolumn{3}{c}{\textbf{ESConv}}\\
    & \textbf{SSR}$\uparrow$ & \textbf{SR}$\uparrow$ & \textbf{AvgT}$\downarrow$ & \textbf{SSR}$\uparrow$ & \textbf{SR}$\uparrow$ & \textbf{AvgT}$\downarrow$\\
    \midrule
    \multicolumn{1}{l}{Standard} & 0.250 & 0.445 & 8.265 & 0.612 & 0.640 & 8.097 \\
    \multicolumn{1}{l}{Proactive~\cite{Deng2023PromptingAE}} & 0.266 & 0.458 & 8.320 & 0.304 & 0.410 & 8.940 \\
    \multicolumn{1}{l}{ProCoT~\cite{Deng2023PromptingAE}} & \underline{0.352} & \underline{0.543} & \underline{7.975} & 0.392 & 0.475 & 8.670 \\
    \multicolumn{1}{l}{ICL-AIF~\cite{Fu2023ImprovingLM}} & 0.301 & 0.465 & 8.085 & 0.709 & 0.744 & \underline{7.429} \\
    \multicolumn{1}{l}{GDP-Zero~\cite{Yu2023PromptBasedMT}} & 0.067 & 0.328 & 9.119 & 0.525 & 0.595 & 8.041 \\
    \multicolumn{1}{l}{PPDPP~\cite{Deng2023PlugandPlayPP}} & 0.266 & 0.463 & 8.185 & 0.664 & 0.709 & 7.652 \\
    \multicolumn{1}{l}{TRIP~\cite{Zhang2024StrengthLI}} & 0.290 & 0.495 & 8.200 & \underline{0.744} & \underline{0.808} & \textbf{6.835} \\
    \midrule
    \midrule
    \multicolumn{1}{l}{UDP (Ours)} & \textbf{0.433} & \textbf{0.598} & \textbf{7.705} & \textbf{0.774} & \textbf{0.832} & 7.585 \\
    \multicolumn{1}{l}{$\quad$-w/o PT} & 0.305 & 0.513 & 8.017 & 0.736 & 0.772 & 7.482 \\
    \multicolumn{1}{l}{$\quad$-w/o RL} & 0.353 & 0.533 & 8.000 & 0.455 & 0.531 & 8.638 \\
    \bottomrule
    \end{tabular}
    }
    \vspace{-0.3cm}
    \label{tab:static_results}
\end{table}

\subsection{Main Results}
We present the evaluation results for all methods under the new protocol in Table~\ref{tab:static_results}. The findings confirm that \textbf{UDP is a robust and effective framework for user-aware dialogue policy planning, capable of addressing diverse user profiles.}
First, UDP consistently outperforms all baselines on both SSR and SR metrics.
Compared to PPDPP, which struggles to learn user-tailored strategies, UDP achieves significant improvements, increasing the SRR and SR by 0.167 and 0.135 on P4G, and by 0.11 and 0.123 on ESConv, respectively.
This demonstrates that UDP has indeed acquired user-tailored policy planning abilities. 
Comparing to TRIP, we achieved improvements of SRR and SR by 0.143 and 0.103 on P4G and 0.03 and 0.024 on ESConv, highlighting its greater advantage in facilitating successful dialogues and improving overall performance across different user types.
However, at the AvgT metric, UDP shows relatively worse performance on ESConv compared with TRIP. This means that UDP needs more dialogue turns on average to therapy the patient.
We find the reason after conducting the human evaluation: compared to TRIP, UDP prefers to comfort the user before offering suggestions, which increases the dialogue rounds significantly. However, this not only facilitates handling the user's issues but also aligns more closely with real scenes. 
Finally, we also conduct ablation studies on the training pipeline. 
Ablation results show that relying solely on pretraining learning (PT) or reinforcement learning (RL) leads to a significant performance drop, which shows the effectiveness of the PT+RL training method.
\begin{table}[!htp]
    \caption{Human evaluation results on ESConv and P4G, with Fleiss Kappa > 0.6.}
    \vspace{-0.3cm}
    \centering
    \renewcommand*{\arraystretch}{1.1}
    \subtable[ESConv]{
    \resizebox{1.0\linewidth}{!}{
    \begin{tabular}{lcccccccc}
    \toprule
    \textbf{UDP} & \multicolumn{2}{c}{\textbf{Ind.}} & \multicolumn{2}{c}{\textbf{Com.}} & \multicolumn{2}{c}{\textbf{Sug.}} & \multicolumn{2}{c}{\textbf{Ove.}} \\
    \cmidrule(lr){2-3}\cmidrule(lr){4-5}\cmidrule(lr){6-7}\cmidrule(lr){8-9}
    vs. & Win & Lose & Win & Lose & Win & Lose & Win & Lose \\
    \midrule
    \textbf{TRIP} & 24.0\% & 13.3\% & 50.7\% & 3.3\% & 10.7\% & 18.7\% & 31.3\% & 19.3\% \\
    \bottomrule
    \end{tabular}
    }
    }
    \subtable[P4G]{
    \resizebox{0.8\linewidth}{!}{
    \begin{tabular}{lcccccc}
    \toprule
    \textbf{UDP} & \multicolumn{2}{c}{\textbf{Inf.}} & \multicolumn{2}{c}{\textbf{Per.}} & \multicolumn{2}{c}{\textbf{Ove.}} \\
    \cmidrule(lr){2-3}\cmidrule(lr){4-5}\cmidrule(lr){6-7}
    vs. & Win & Lose & Win & Lose & Win & Lose \\
    \midrule
    \textbf{TRIP} & 22.5\% & 18.8\% & 28.3\% & 14.6\% & 20.8\% & 10.4\% \\
    \bottomrule
    \end{tabular}
    }
    }
    \vspace{-0.5cm}
    \label{tab:human_eval}
\end{table}

\subsection{Human Evaluation}
Following prior work~\cite{Deng2023PlugandPlayPP}, we conduct Human Evaluation on generated responses. We randomly selected 50 test samples for ESConv and entire 80 test samples for P4G. Three annotators compare the generated responses by UDP and TRIP on ESConv from four perspectives: Suggestion (\textbf{Sug.}, comparing the quality of suggestions), Identification (\textbf{Ide.}, comparing the proactivity in asking about the emotional issue), Comforting (\textbf{Com.}, comparing the quality of comfort), and Overall (\textbf{Ove.}), 
and from three perspectives on P4G: Informative (\textbf{Inf.}, comparing whether the provided information is detailed), Persuative (\textbf{Per.}, comparing the persuasiveness of responses), and Overall (\textbf{Ove.}).
Results are shown in Table~\ref{tab:human_eval}.

Firstly, UDP outperforms TRIP in all aspects on P4G, aligning with automatic results in Table~\ref{tab:static_results}. 
On ESConv, UDP exhibits a clear advantage in empathy. 
This can be attributed to the fact that UDP has learned that first providing comfort to stabilize the user's emotions before offering suggestions effectively addresses most cases facing different user types. This pipeline also aligns with the real human emotion support process.
In contrast, TRIP tends to give suggestions more quickly, which may be rejected when the user’s emotional state has not been fully stabilized.
While UDP performs slightly below TRIP (10.7\% vs. 18.7\%) on suggesting, this does not imply UDP fails to choose the ``Suggestion'' action during policy planning in these cases; rather, the quality of suggestions generated under UDP's guidance is slightly inferior to those generated by TRIP in terms of usefulness.
The reason may be that UDP spends many turns to comfort the user before suggestions, leaving fewer turns for thinking more valuable suggestions without sufficient user feedback (we set the maximum number of turns as 10).

\begin{figure}[!t]
    \centering
    \subfigure[SSR on ESConv]{
        \includegraphics[width=0.475\linewidth]{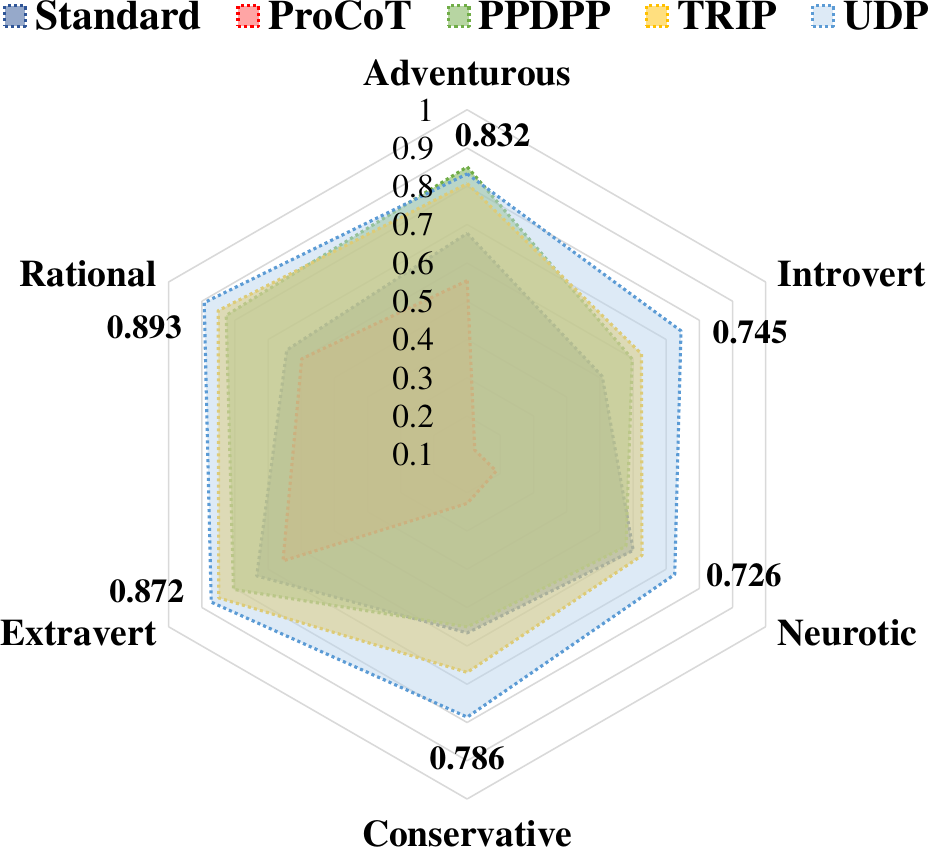}
    }
    \subfigure[SR on ESConv]{
        \includegraphics[width=0.475\linewidth]{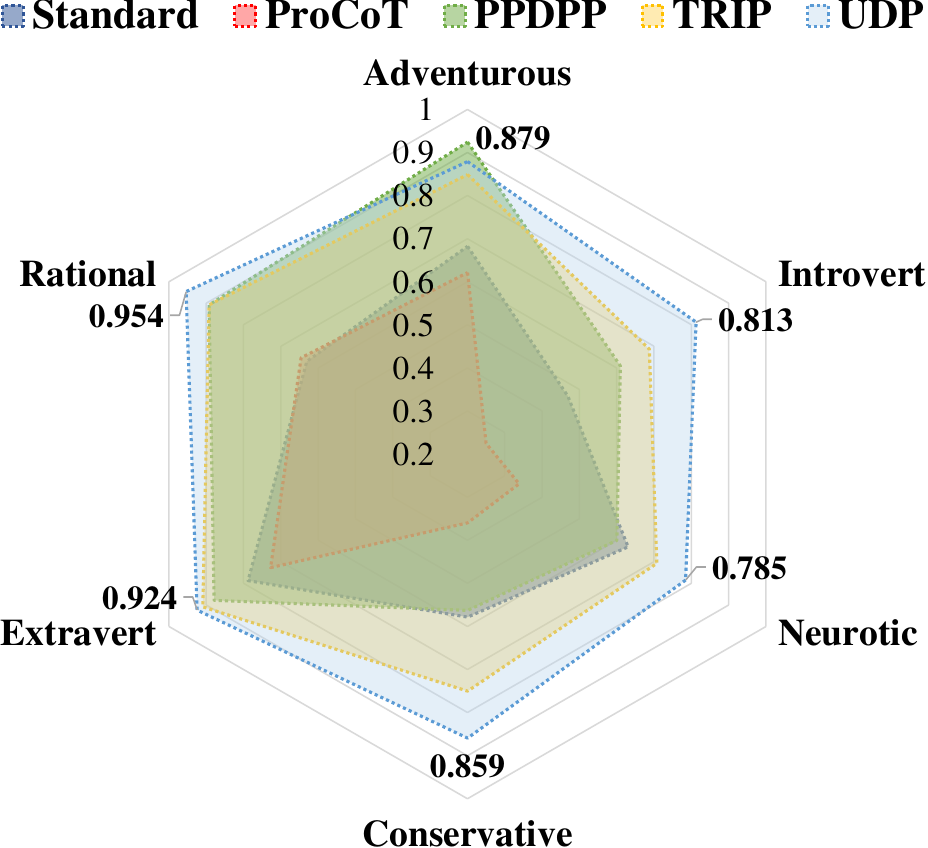}
    }
    \subfigure[SSR on P4G]{
        \includegraphics[width=0.475\linewidth]{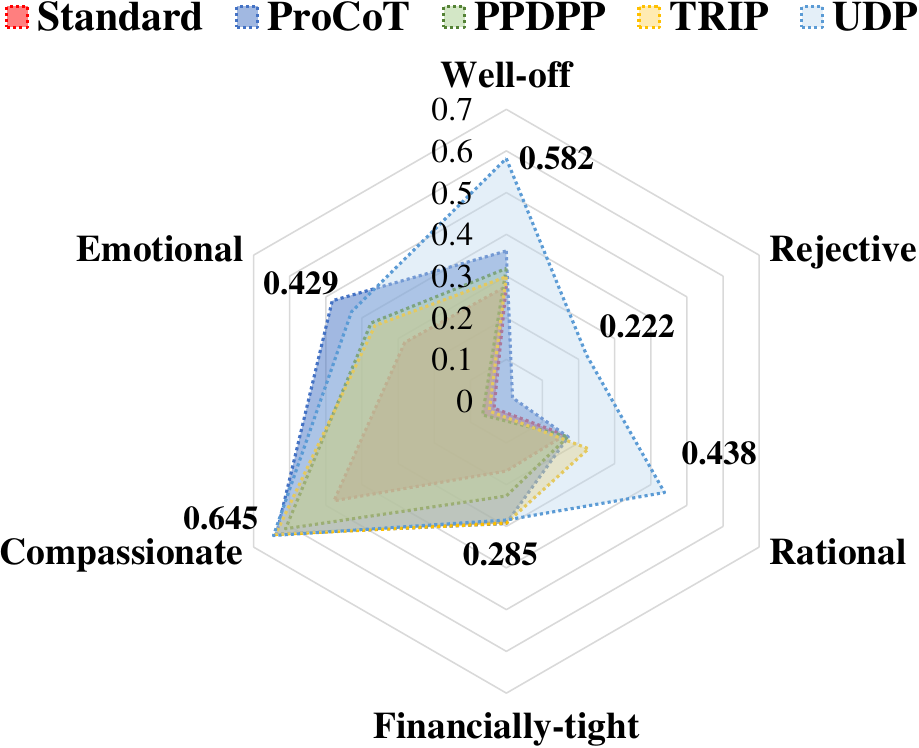}
    }
    \subfigure[SR on P4G]{
        \includegraphics[width=0.475\linewidth]{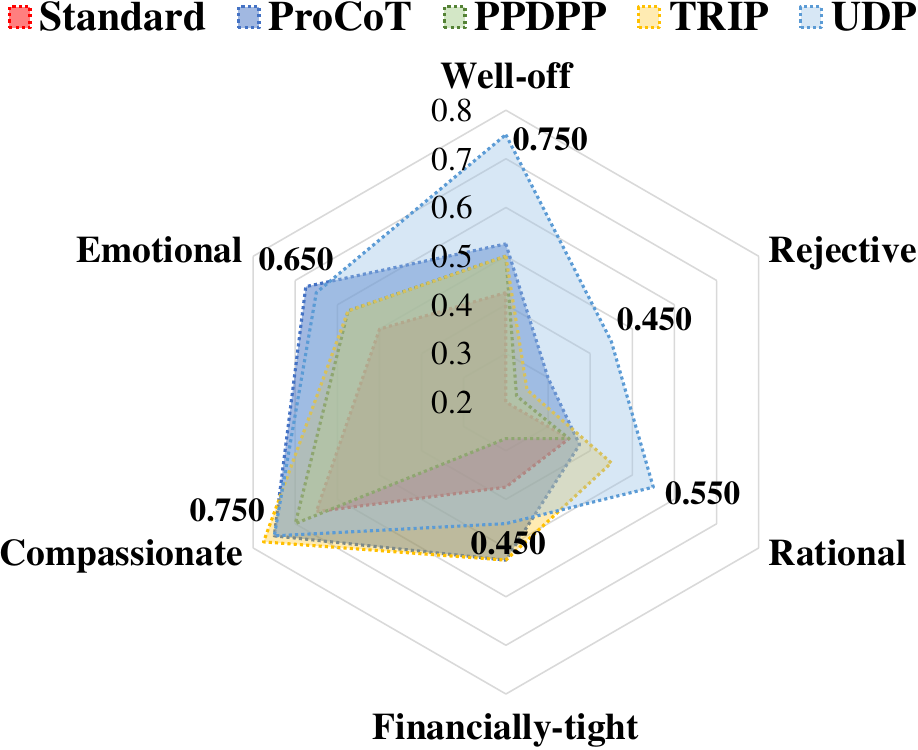}
    }
    \vspace{-0.3cm}
    \caption{The dialogue agents' performance across various personas on P4G and ESConv. UDP achieves improvements on almost all personas.}
    \vspace{-0.3cm}
    \label{fig:persona_analysis}
\end{figure}

\subsection{Performance Across Personas}
To investigate how UDP enhances the agent's policy planning capabilities, we conducted a detailed analysis based on specific user personas. The SSR and SR metrics were evaluated separately for all user personas across different datasets, with results presented in Figure~\ref{fig:persona_analysis}.
Firstly, UDP outperforms baselines across most user types on ESConv. Specifically, UDP not only obtains improvements on the easy ``Rational'' users, but also achieves significant gains with more challenging user types, including ``Introvert'', ``Neurotic'', and ``Conservative''. Compared to TRIP, UDP improves the SR metric by 0.125, 0.769, and 0.109 on these three personas, respectively. This can be attributed to UDP's ability to adapt strategies according to specific user types and the introduction of Active Learning-based optimization, which focuses more attention on these difficult user types during training. Similarly, on P4G, UDP demonstrates significant improvements on user types where other baselines perform poorly, such as ``Well-off'', ``Rejective'', and ``Rational'', with SR improvements of 0.25, 0.2, and 0.1 compared to TRIP. These improvements further validate UDP’s effectiveness. However, UDP shows slight declines on ``Emotional'' and ``Financially-tight'' user types. The reason may be UDP's overemphasis on other user types during the active learning phase. Active learning with more effective and balanced methods will be explored in future work.

\begin{table}[!htp]
    \caption{Ablation study results on P4G and ESConv. ``AL'' means ``Active Learning''. ``S1'' refers to ``Stage 1'' and means portraying the user profile; ``S2'' refers to ``Stage 2'' and means anticipating the user feedback before planning.}
    \vspace{-0.3cm}
    \centering
    \renewcommand*{\arraystretch}{1.1}
    \subtable[P4G]{
    \resizebox{0.47\linewidth}{!}{
    \begin{tabular}{ccc}
    \toprule
    \multirow{2}{*}{\textbf{Models}} & \multicolumn{2}{c}{\textbf{P4G}} \\
    & \textbf{SSR} & \textbf{SR} \\
    \midrule
    \multicolumn{1}{c}{UDP} & \textbf{0.433} & \textbf{0.598} \\
    \multicolumn{1}{c}{-w/o AL} & 0.392 & 0.563 \\
    \multicolumn{1}{c}{-w/o S2} & 0.344 & 0.525 \\
    \multicolumn{1}{c}{-w/o S1\&2} & 0.322 & 0.500 \\
    \bottomrule
    \end{tabular}
    }
    }
    \subtable[ESConv]{
    \resizebox{0.47\linewidth}{!}{
    \begin{tabular}{ccc}
    \toprule
    \multirow{2}{*}{\textbf{Models}} & \multicolumn{2}{c}{\textbf{ESConv}}\\
    & \textbf{SSR} & \textbf{SR} \\
    \midrule
    \multicolumn{1}{c}{UDP} & 0.742 & 0.791 \\
    \multicolumn{1}{c}{-w/o AL} & 0.729 & 0.774 \\
    \multicolumn{1}{c}{-w/o S2} & \textbf{0.774} & \textbf{0.832} \\
    \multicolumn{1}{c}{-w/o S1\&2} & 0.713 & 0.773 \\
    \bottomrule
    \end{tabular}
    }
    }
    \vspace{-0.4cm}
    \label{tab:ablation_results}
\end{table}

\subsection{Ablation Study}
To investigate the contributions of different modules and evaluate the effectiveness of our proposed active learning-based techniques, we conducted ablation studies. Specifically, we removed Stage 2, Stage 1\&2, and the active learning-based user profile sampling method. Removing the 1st stage alone was not considered because the 2nd stage requires the user profile types inferred in the 1st stage as input. 
The final results are summarized in Table~\ref{tab:ablation_results}. On P4G, we observe that the complete framework performs the best. Removing the active learning training results in a slight performance drop. Progressively eliminating Stage 2 and Stage 1 leads to further degradation in SSR metrics, which aligns with our hypothesis. 
On ESConv, the ``w/o Stage 2'' setting surpasses the complete framework and achieves the best performance. We attribute this to the differences in dialogue tasks: P4G involves non-cooperative dialogues where anticipating user feedback and tailoring responses are crucial for effective communication. In contrast, ESConv represents cooperative dialogues where the therapist focuses more on selecting appropriate strategies based on patients' personalities and preferences. 
Besides, in the emotion support task, user reactions become highly unpredictable due to emotional instability (e.g., ``Neurotic'' users), which further increases behavior uncertainty.

\begin{figure}[t]
    \centering
    \subfigure[P4G]{
        \includegraphics[width=0.475\linewidth]{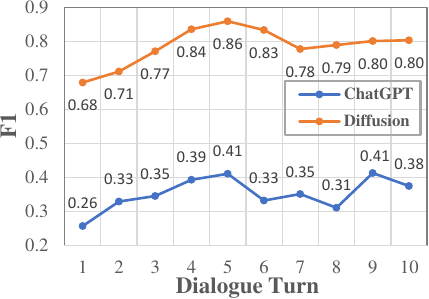}
    }
    \subfigure[ESConv]{
        \includegraphics[width=0.475\linewidth]{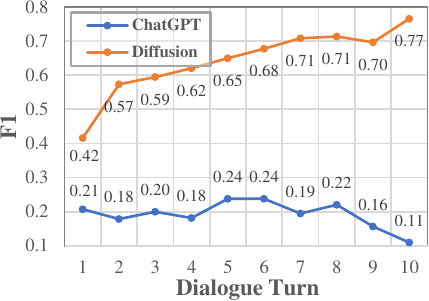}
    }
    \vspace{-0.3cm}
    \caption{Changes in user persona prediction accuracy with increasing dialogue turns. 
    Compared to ChatGPT, our proposed diffusion-based Portrayer performs better.
    }
    \vspace{-0.2cm}\label{fig:persona_pred_analysis}
\end{figure}

\subsection{More Analyses}
\subsubsection{Diffusion-based Profile Portrayer Analysis}
To evaluate our proposed Profile Portrayer, we compare it with the method that uses an LLM to infer user persona categories. 
During the pretraining phase of Stage 1, we evaluate the Diffusion Model-based User Profile Portrayer on the valid set to select the best pretrained Portrayer model. 
To compare with LLMs, we replace the diffusion model with ChatGPT to infer the user types directly after each dialogue turn. By comparing the F1 scores of user persona prediction between these two Portrayer, we hope to verify the effectiveness of our proposed method.
The experimental results, shown in Figure~\ref{fig:persona_pred_analysis}, clearly demonstrate that the Diffusion Model-based User Profile Portrayer significantly outperforms ChatGPT. 
Moreover, as dialogues progress, the F1 score of the Diffusion Model shows a notable upward trend, particularly in ESConv, indicating that user profile predictions become more accurate over time—aligning with intuitive expectations. 
In contrast, ChatGPT shows no such trend and even exhibits a decline in ESConv. 
This decline likely stems from the fact that user profiles in ESConv are designed to emphasize internal characteristics, requiring advanced comprehension, analysis, and reasoning capabilities from the LLM.
However, our proposed Diffusion-based Portrayer is able to learn the aforementioned capabilities through targeted training, thus demonstrating superior user-type prediction performance. While training a LLM may achieve the same, even better performances, it requires expensive computational resources. In contrast, our lightweight Portrayer achieves excellent performances with fewer training costs. 

\begin{table}[t]
    \caption{The strategy distribution difference results. 
    }
    \vspace{-0.3cm}
    \centering
    \renewcommand*{\arraystretch}{1.1}
    \subtable[P4G]{
    \resizebox{0.42\linewidth}{!}{
    \begin{tabular}{lcc}
    \toprule
    \textbf{Models} & \textbf{Intra}$\downarrow$ & \textbf{Inter}$\uparrow$ \\
    \midrule
    \multicolumn{1}{l}{ProCoT} & 0.105 & 0.118 \\
    \multicolumn{1}{l}{PPDPP} & 0.146 & 0.212 \\
    \multicolumn{1}{l}{TRIP} & 0.078 & 0.033 \\
    \midrule
    \multicolumn{1}{l}{\textbf{UDP}} & 0.052 & 0.193 \\
    \bottomrule
    \end{tabular}
    }
    }
    \subtable[ESConv]{
    \resizebox{0.42\linewidth}{!}{
    \begin{tabular}{lcc}
    \toprule
    \textbf{Models} & \textbf{Intra}$\downarrow$ & \textbf{Inter}$\uparrow$ \\
    \midrule
    \multicolumn{1}{l}{ProCoT} & 0.010 & 0.056 \\
    \multicolumn{1}{l}{PPDPP} & 0.028 & 0.022 \\
    \multicolumn{1}{l}{TRIP} & 0.043 & 0.138 \\
    \midrule
    \multicolumn{1}{l}{\textbf{UDP}} & 0.030 & 0.163 \\
    \bottomrule
    \end{tabular}
    }
    }
    \vspace{-0.3cm}
    \label{tab:strategy_distribution}
\end{table}

\subsubsection{User-Aware Strategy Distribution Analysis}
We analyze the impact of user personas on strategy selection by calculating the differences in strategies across examples. Specifically, we normalize strategy usage frequencies into distributions and compute two metrics: \textbf{Intra} (average JS divergence~\cite{menendez1997jensen} between distributions for different cases under the same persona) and \textbf{Inter} (average JS divergence between distributions for different personas). Results in Table~\ref{tab:strategy_distribution} show key results.
UDP exhibits significant differences in Intra \textit{vs.} Inter on P4G (0.052 \textit{vs.} 0.193) and ESConv (0.030 \textit{vs.} 0.163), demonstrating its ability to adopt persona-specific strategies and plan user-tailored responses. Conversely, ProCoT on P4G (0.105 \textit{vs.} 0.118) and PPDPP on ESConv (0.028 \textit{vs.} 0.022) show similar Intra \textit{vs.} Inter values, indicating limited adaptability to user personas.


\section{Conclusion}
In this study, we explored effective dialogue policy planning tailored to diverse user characteristics. We developed task-specific user personas to simulate realistic interaction scenarios and conducted a comprehensive evaluation to assess the limitations of existing dialogue agents. Based on these insights, we proposed User-Tailored Dialogue Policy Planning (UDP), a novel framework that incorporates an Intrinsic User World Model to dynamically adjust strategies based on evolving user traits.
UDP consists of three key components: a diffusion-based user persona portrayer for dynamic user profiling, a Brownian Bridge Process-based user feedback anticipator to predict responses, and a user-tailored policy planner to optimize dialogue strategies. Extensive experiments across diverse tasks validate the effectiveness and adaptability of our framework, demonstrating significant improvements in user-specific dialogue planning.
We believe that UDP provides a promising foundation for advancing personalized dialogue policy planning, offering a more adaptive and human-centric approach to conversational AI.

\section{Acknowledgments}
The research in this article is supported by the National Science Foundation of China (U22B2059, 62276083). We also appreciate the support from China Mobile Group Heilongjiang Co., Ltd. @ on our research; the research is jointly completed by both parties. This research is also supported by the Ministry of Education, Singapore, under its AcRF Tier 2 Funding (Proposal ID: T2EP20123-0052). Any opinions, findings and conclusions or recommendations expressed in this material are those of the author(s) and do not reflect the views of the Ministry of Education, Singapore.


\clearpage
\bibliographystyle{ACM-Reference-Format}
\balance
\bibliography{ref}


\begin{thebibliography}{54}


\ifx \showCODEN    \undefined \def \showCODEN     #1{\unskip}     \fi
\ifx \showDOI      \undefined \def \showDOI       #1{#1}\fi
\ifx \showISBNx    \undefined \def \showISBNx     #1{\unskip}     \fi
\ifx \showISBNxiii \undefined \def \showISBNxiii  #1{\unskip}     \fi
\ifx \showISSN     \undefined \def \showISSN      #1{\unskip}     \fi
\ifx \showLCCN     \undefined \def \showLCCN      #1{\unskip}     \fi
\ifx \shownote     \undefined \def \shownote      #1{#1}          \fi
\ifx \showarticletitle \undefined \def \showarticletitle #1{#1}   \fi
\ifx \showURL      \undefined \def \showURL       {\relax}        \fi
\providecommand\bibfield[2]{#2}
\providecommand\bibinfo[2]{#2}
\providecommand\natexlab[1]{#1}
\providecommand\showeprint[2][]{arXiv:#2}

\bibitem[Asri et~al\mbox{.}(2016)]%
        {Asri2016ASM}
\bibfield{author}{\bibinfo{person}{Layla~El Asri}, \bibinfo{person}{Jing He}, {and} \bibinfo{person}{Kaheer Suleman}.} \bibinfo{year}{2016}\natexlab{}.
\newblock \showarticletitle{A Sequence-to-Sequence Model for User Simulation in Spoken Dialogue Systems}. In \bibinfo{booktitle}{\emph{Interspeech}}.
\newblock
\urldef\tempurl%
\url{https://arxiv.org/pdf/1607.00070}
\showURL{%
\tempurl}


\bibitem[Bang et~al\mbox{.}(2023)]%
        {Bang2023AMM}
\bibfield{author}{\bibinfo{person}{Yejin Bang}, \bibinfo{person}{Samuel Cahyawijaya}, \bibinfo{person}{Nayeon Lee}, \bibinfo{person}{Wenliang Dai}, \bibinfo{person}{Dan Su}, \bibinfo{person}{Bryan Wilie}, \bibinfo{person}{Holy Lovenia}, \bibinfo{person}{Ziwei Ji}, \bibinfo{person}{Tiezheng Yu}, \bibinfo{person}{Willy Chung}, \bibinfo{person}{Quyet~V. Do}, \bibinfo{person}{Yan Xu}, {and} \bibinfo{person}{Pascale Fung}.} \bibinfo{year}{2023}\natexlab{}.
\newblock \showarticletitle{A Multitask, Multilingual, Multimodal Evaluation of ChatGPT on Reasoning, Hallucination, and Interactivity}.
\newblock \bibinfo{journal}{\emph{ArXiv}}  \bibinfo{volume}{abs/2302.04023} (\bibinfo{year}{2023}).
\newblock
\urldef\tempurl%
\url{https://aclanthology.org/2023.ijcnlp-main.45.pdf}
\showURL{%
\tempurl}


\bibitem[Chen et~al\mbox{.}(2024b)]%
        {Chen2024FromPT}
\bibfield{author}{\bibinfo{person}{Jiangjie Chen}, \bibinfo{person}{Xintao Wang}, \bibinfo{person}{Rui Xu}, \bibinfo{person}{Siyu Yuan}, \bibinfo{person}{Yikai Zhang}, \bibinfo{person}{Wei Shi}, \bibinfo{person}{Jian Xie}, \bibinfo{person}{Shuang Li}, \bibinfo{person}{Ruihan Yang}, \bibinfo{person}{Tinghui Zhu}, \bibinfo{person}{Aili Chen}, \bibinfo{person}{Nianqi Li}, \bibinfo{person}{Lida Chen}, \bibinfo{person}{Caiyu Hu}, \bibinfo{person}{Siye Wu}, \bibinfo{person}{Scott Ren}, \bibinfo{person}{Ziquan Fu}, {and} \bibinfo{person}{Yanghua Xiao}.} \bibinfo{year}{2024}\natexlab{b}.
\newblock \showarticletitle{From Persona to Personalization: A Survey on Role-Playing Language Agents}.
\newblock \bibinfo{journal}{\emph{ArXiv}}  \bibinfo{volume}{abs/2404.18231} (\bibinfo{year}{2024}).
\newblock
\urldef\tempurl%
\url{https://arxiv.org/pdf/2404.18231}
\showURL{%
\tempurl}


\bibitem[Chen et~al\mbox{.}(2024a)]%
        {Chen2024TheOO}
\bibfield{author}{\bibinfo{person}{Nuo Chen}, \bibinfo{person}{Yang Deng}, {and} \bibinfo{person}{Jia Li}.} \bibinfo{year}{2024}\natexlab{a}.
\newblock \showarticletitle{The Oscars of AI Theater: A Survey on Role-Playing with Language Models}.
\newblock \bibinfo{journal}{\emph{ArXiv}}  \bibinfo{volume}{abs/2407.11484} (\bibinfo{year}{2024}).
\newblock
\urldef\tempurl%
\url{https://arxiv.org/pdf/2407.11484}
\showURL{%
\tempurl}


\bibitem[chin Lin et~al\mbox{.}(2023)]%
        {Lin2023EmoUSSU}
\bibfield{author}{\bibinfo{person}{Hsien chin Lin}, \bibinfo{person}{Shutong Feng}, \bibinfo{person}{Christian Geishauser}, \bibinfo{person}{Nurul Lubis}, \bibinfo{person}{Carel van Niekerk}, \bibinfo{person}{Michael Heck}, \bibinfo{person}{Benjamin~Matthias Ruppik}, \bibinfo{person}{Renato Vukovic}, {and} \bibinfo{person}{Milica Gavsi'c}.} \bibinfo{year}{2023}\natexlab{}.
\newblock \showarticletitle{EmoUS: Simulating User Emotions in Task-Oriented Dialogues}.
\newblock \bibinfo{journal}{\emph{Proceedings of the 46th International ACM SIGIR Conference on Research and Development in Information Retrieval}} (\bibinfo{year}{2023}).
\newblock
\urldef\tempurl%
\url{https://dl.acm.org/doi/10.1145/3539618.3592092}
\showURL{%
\tempurl}


\bibitem[Christakopoulou et~al\mbox{.}(2018)]%
        {christakopoulou2018q}
\bibfield{author}{\bibinfo{person}{Konstantina Christakopoulou}, \bibinfo{person}{Alex Beutel}, \bibinfo{person}{Rui Li}, \bibinfo{person}{Sagar Jain}, {and} \bibinfo{person}{Ed~H Chi}.} \bibinfo{year}{2018}\natexlab{}.
\newblock \showarticletitle{Q\&R: A two-stage approach toward interactive recommendation}. In \bibinfo{booktitle}{\emph{Proceedings of the 24th ACM SIGKDD international conference on knowledge discovery \& data mining}}. \bibinfo{pages}{139--148}.
\newblock


\bibitem[Dao et~al\mbox{.}(2024)]%
        {Dao2024ExperienceAS}
\bibfield{author}{\bibinfo{person}{Huy Dao}, \bibinfo{person}{Yang Deng}, \bibinfo{person}{Khanh-Huyen Bui}, \bibinfo{person}{Dung~D. Le}, {and} \bibinfo{person}{Lizi Liao}.} \bibinfo{year}{2024}\natexlab{}.
\newblock \showarticletitle{Experience as Source for Anticipation and Planning: Experiential Policy Learning for Target-driven Recommendation Dialogues}. In \bibinfo{booktitle}{\emph{Conference on Empirical Methods in Natural Language Processing}}.
\newblock
\urldef\tempurl%
\url{https://api.semanticscholar.org/CorpusID:273822277}
\showURL{%
\tempurl}


\bibitem[Deng et~al\mbox{.}(2023a)]%
        {Deng2023PromptingAE}
\bibfield{author}{\bibinfo{person}{Yang Deng}, \bibinfo{person}{Wenqiang Lei}, \bibinfo{person}{Hongru Wang}, {and} \bibinfo{person}{Tat seng Chua}.} \bibinfo{year}{2023}\natexlab{a}.
\newblock \showarticletitle{Prompting and Evaluating Large Language Models for Proactive Dialogues: Clarification, Target-guided, and Non-collaboration}. In \bibinfo{booktitle}{\emph{Conference on Empirical Methods in Natural Language Processing}}.
\newblock
\urldef\tempurl%
\url{https://arxiv.org/pdf/2305.13626}
\showURL{%
\tempurl}


\bibitem[Deng et~al\mbox{.}(2023b)]%
        {Deng2023PlugandPlayPP}
\bibfield{author}{\bibinfo{person}{Yang Deng}, \bibinfo{person}{Wenxuan Zhang}, \bibinfo{person}{Wai Lam}, \bibinfo{person}{See-Kiong Ng}, {and} \bibinfo{person}{Tat-Seng Chua}.} \bibinfo{year}{2023}\natexlab{b}.
\newblock \showarticletitle{Plug-and-Play Policy Planner for Large Language Model Powered Dialogue Agents}.
\newblock \bibinfo{journal}{\emph{ArXiv}}  \bibinfo{volume}{abs/2311.00262} (\bibinfo{year}{2023}).
\newblock
\urldef\tempurl%
\url{https://arxiv.org/pdf/2311.00262}
\showURL{%
\tempurl}


\bibitem[Dhariwal and Nichol(2021)]%
        {Dhariwal2021DiffusionMB}
\bibfield{author}{\bibinfo{person}{Prafulla Dhariwal} {and} \bibinfo{person}{Alex Nichol}.} \bibinfo{year}{2021}\natexlab{}.
\newblock \showarticletitle{Diffusion Models Beat GANs on Image Synthesis}.
\newblock \bibinfo{journal}{\emph{ArXiv}}  \bibinfo{volume}{abs/2105.05233} (\bibinfo{year}{2021}).
\newblock
\urldef\tempurl%
\url{https://arxiv.org/pdf/2105.05233}
\showURL{%
\tempurl}


\bibitem[Ding et~al\mbox{.}(2024)]%
        {Ding2024UnderstandingWO}
\bibfield{author}{\bibinfo{person}{Jingtao Ding}, \bibinfo{person}{Yunke Zhang}, \bibinfo{person}{Yu Shang}, \bibinfo{person}{Yuheng Zhang}, \bibinfo{person}{Zefang Zong}, \bibinfo{person}{Jie Feng}, \bibinfo{person}{Yuan Yuan}, \bibinfo{person}{Hongyuan Su}, \bibinfo{person}{Nian Li}, \bibinfo{person}{Nicholas Sukiennik}, \bibinfo{person}{Fengli Xu}, {and} \bibinfo{person}{Yong Li}.} \bibinfo{year}{2024}\natexlab{}.
\newblock \showarticletitle{Understanding World or Predicting Future? A Comprehensive Survey of World Models}.
\newblock
\urldef\tempurl%
\url{https://arxiv.org/pdf/2411.14499}
\showURL{%
\tempurl}


\bibitem[Fu et~al\mbox{.}(2023a)]%
        {fu2023improving}
\bibfield{author}{\bibinfo{person}{Yao Fu}, \bibinfo{person}{Hao Peng}, \bibinfo{person}{Tushar Khot}, {and} \bibinfo{person}{Mirella Lapata}.} \bibinfo{year}{2023}\natexlab{a}.
\newblock \showarticletitle{Improving language model negotiation with self-play and in-context learning from ai feedback}.
\newblock \bibinfo{journal}{\emph{arXiv preprint arXiv:2305.10142}} (\bibinfo{year}{2023}).
\newblock


\bibitem[Fu et~al\mbox{.}(2023b)]%
        {Fu2023ImprovingLM}
\bibfield{author}{\bibinfo{person}{Yao Fu}, \bibinfo{person}{Hao-Chun Peng}, \bibinfo{person}{Tushar Khot}, {and} \bibinfo{person}{Mirella Lapata}.} \bibinfo{year}{2023}\natexlab{b}.
\newblock \showarticletitle{Improving Language Model Negotiation with Self-Play and In-Context Learning from AI Feedback}.
\newblock \bibinfo{journal}{\emph{ArXiv}}  \bibinfo{volume}{abs/2305.10142} (\bibinfo{year}{2023}).
\newblock
\urldef\tempurl%
\url{https://arxiv.org/pdf/2305.10142}
\showURL{%
\tempurl}


\bibitem[Gao et~al\mbox{.}(2018)]%
        {Gao2018NeuralAT}
\bibfield{author}{\bibinfo{person}{Jianfeng Gao}, \bibinfo{person}{Michel Galley}, {and} \bibinfo{person}{Lihong Li}.} \bibinfo{year}{2018}\natexlab{}.
\newblock \showarticletitle{Neural Approaches to Conversational AI}.
\newblock \bibinfo{journal}{\emph{The 41st International ACM SIGIR Conference on Research \& Development in Information Retrieval}} (\bibinfo{year}{2018}).
\newblock
\urldef\tempurl%
\url{https://dl.acm.org/doi/10.1145/3209978.3210183}
\showURL{%
\tempurl}


\bibitem[Goldberg(1992)]%
        {Goldberg1992THEDO}
\bibfield{author}{\bibinfo{person}{Lewis~R. Goldberg}.} \bibinfo{year}{1992}\natexlab{}.
\newblock \showarticletitle{THE DEVELOPMENT OF MARKERS FOR THE BIG-FIVE FACTOR STRUCTURE}.
\newblock \bibinfo{journal}{\emph{Psychological Assessment}}  \bibinfo{volume}{4} (\bibinfo{year}{1992}), \bibinfo{pages}{26--42}.
\newblock
\urldef\tempurl%
\url{https://psycnet.apa.org/doiLanding?doi=10.1037/1040-3590.4.1.26}
\showURL{%
\tempurl}


\bibitem[Guo et~al\mbox{.}(2024)]%
        {Guo2024PCQPRPC}
\bibfield{author}{\bibinfo{person}{Shasha Guo}, \bibinfo{person}{Lizi Liao}, \bibinfo{person}{Jing Zhang}, \bibinfo{person}{Cuiping Li}, {and} \bibinfo{person}{Hong Chen}.} \bibinfo{year}{2024}\natexlab{}.
\newblock \showarticletitle{PCQPR: Proactive Conversational Question Planning with Reflection}. In \bibinfo{booktitle}{\emph{Conference on Empirical Methods in Natural Language Processing}}.
\newblock
\urldef\tempurl%
\url{https://aclanthology.org/2024.emnlp-main.631.pdf}
\showURL{%
\tempurl}


\bibitem[Ha and Schmidhuber(2018)]%
        {Ha2018WorldM}
\bibfield{author}{\bibinfo{person}{David~R Ha} {and} \bibinfo{person}{J{\"u}rgen Schmidhuber}.} \bibinfo{year}{2018}\natexlab{}.
\newblock \showarticletitle{World Models}.
\newblock \bibinfo{journal}{\emph{ArXiv}}  \bibinfo{volume}{abs/1803.10122} (\bibinfo{year}{2018}).
\newblock
\urldef\tempurl%
\url{https://arxiv.org/pdf/1803.10122}
\showURL{%
\tempurl}


\bibitem[He et~al\mbox{.}(2024a)]%
        {He2024PlanningLH}
\bibfield{author}{\bibinfo{person}{Tao He}, \bibinfo{person}{Lizi Liao}, \bibinfo{person}{Yixin Cao}, \bibinfo{person}{Yuanxing Liu}, \bibinfo{person}{Ming Liu}, \bibinfo{person}{Zerui Chen}, {and} \bibinfo{person}{Bing Qin}.} \bibinfo{year}{2024}\natexlab{a}.
\newblock \showarticletitle{Planning Like Human: A Dual-process Framework for Dialogue Planning}.
\newblock \bibinfo{journal}{\emph{ArXiv}}  \bibinfo{volume}{abs/2406.05374} (\bibinfo{year}{2024}).
\newblock
\urldef\tempurl%
\url{https://arxiv.org/pdf/2406.05374}
\showURL{%
\tempurl}


\bibitem[He et~al\mbox{.}(2024b)]%
        {He2024SimulationFreeHL}
\bibfield{author}{\bibinfo{person}{Tao He}, \bibinfo{person}{Lizi Liao}, \bibinfo{person}{Yixin Cao}, \bibinfo{person}{Yuanxing Liu}, \bibinfo{person}{Yiheng Sun}, \bibinfo{person}{Zerui Chen}, \bibinfo{person}{Ming Liu}, {and} \bibinfo{person}{Bing Qin}.} \bibinfo{year}{2024}\natexlab{b}.
\newblock \showarticletitle{Simulation-Free Hierarchical Latent Policy Planning for Proactive Dialogues}.
\newblock
\urldef\tempurl%
\url{https://arxiv.org/pdf/2412.14584}
\showURL{%
\tempurl}


\bibitem[Ho et~al\mbox{.}(2020)]%
        {Ho2020DenoisingDP}
\bibfield{author}{\bibinfo{person}{Jonathan Ho}, \bibinfo{person}{Ajay Jain}, {and} \bibinfo{person}{P. Abbeel}.} \bibinfo{year}{2020}\natexlab{}.
\newblock \showarticletitle{Denoising Diffusion Probabilistic Models}.
\newblock \bibinfo{journal}{\emph{ArXiv}}  \bibinfo{volume}{abs/2006.11239} (\bibinfo{year}{2020}).
\newblock
\urldef\tempurl%
\url{https://arxiv.org/pdf/2006.11239}
\showURL{%
\tempurl}


\bibitem[Kaiser et~al\mbox{.}(2019)]%
        {Kaiser2019ModelBasedRL}
\bibfield{author}{\bibinfo{person}{Lukasz Kaiser}, \bibinfo{person}{Mohammad Babaeizadeh}, \bibinfo{person}{Piotr Milos}, \bibinfo{person}{Blazej Osinski}, \bibinfo{person}{Roy~H. Campbell}, \bibinfo{person}{K. Czechowski}, \bibinfo{person}{D. Erhan}, \bibinfo{person}{Chelsea Finn}, \bibinfo{person}{Piotr Kozakowski}, \bibinfo{person}{Sergey Levine}, \bibinfo{person}{Afroz Mohiuddin}, \bibinfo{person}{Ryan Sepassi}, \bibinfo{person}{G. Tucker}, {and} \bibinfo{person}{Henryk Michalewski}.} \bibinfo{year}{2019}\natexlab{}.
\newblock \showarticletitle{Model-Based Reinforcement Learning for Atari}.
\newblock \bibinfo{journal}{\emph{ArXiv}}  \bibinfo{volume}{abs/1903.00374} (\bibinfo{year}{2019}).
\newblock
\urldef\tempurl%
\url{https://arxiv.org/pdf/1903.00374}
\showURL{%
\tempurl}


\bibitem[Kwan et~al\mbox{.}(2022)]%
        {Kwan2022ASO}
\bibfield{author}{\bibinfo{person}{Wai-Chung Kwan}, \bibinfo{person}{Hongru Wang}, \bibinfo{person}{Huimin Wang}, {and} \bibinfo{person}{Kam-Fai Wong}.} \bibinfo{year}{2022}\natexlab{}.
\newblock \showarticletitle{A Survey on Recent Advances and Challenges in Reinforcement Learning Methods for Task-oriented Dialogue Policy Learning}.
\newblock \bibinfo{journal}{\emph{Machine Intelligence Research}} (\bibinfo{year}{2022}), \bibinfo{pages}{1--17}.
\newblock
\urldef\tempurl%
\url{https://link.springer.com/article/10.1007/s11633-022-1347-y}
\showURL{%
\tempurl}


\bibitem[LeCun and Courant(2022)]%
        {LeCun2022APT}
\bibfield{author}{\bibinfo{person}{Yann LeCun} {and} \bibinfo{person}{Courant}.} \bibinfo{year}{2022}\natexlab{}.
\newblock \showarticletitle{A Path Towards Autonomous Machine Intelligence Version 0.9.2, 2022-06-27}.
\newblock
\urldef\tempurl%
\url{https://api.semanticscholar.org/CorpusID:251881108}
\showURL{%
\tempurl}


\bibitem[Li et~al\mbox{.}(2023)]%
        {Li2023CAMELCA}
\bibfield{author}{\bibinfo{person}{G. Li}, \bibinfo{person}{Hasan Hammoud}, \bibinfo{person}{Hani Itani}, \bibinfo{person}{Dmitrii Khizbullin}, {and} \bibinfo{person}{Bernard Ghanem}.} \bibinfo{year}{2023}\natexlab{}.
\newblock \showarticletitle{CAMEL: Communicative Agents for "Mind" Exploration of Large Language Model Society}. In \bibinfo{booktitle}{\emph{Neural Information Processing Systems}}.
\newblock
\urldef\tempurl%
\url{https://arxiv.org/pdf/2303.17760}
\showURL{%
\tempurl}


\bibitem[Li et~al\mbox{.}(2014)]%
        {Li2014TemporalSL}
\bibfield{author}{\bibinfo{person}{Lihong Li}, \bibinfo{person}{He He}, {and} \bibinfo{person}{J. Williams}.} \bibinfo{year}{2014}\natexlab{}.
\newblock \showarticletitle{Temporal supervised learning for inferring a dialog policy from example conversations}.
\newblock \bibinfo{journal}{\emph{2014 IEEE Spoken Language Technology Workshop (SLT)}} (\bibinfo{year}{2014}), \bibinfo{pages}{312--317}.
\newblock
\urldef\tempurl%
\url{https://ieeexplore.ieee.org/document/7078593/}
\showURL{%
\tempurl}


\bibitem[Liang et~al\mbox{.}(2024)]%
        {Liang2024ActivelyLF}
\bibfield{author}{\bibinfo{person}{Jinggui Liang}, \bibinfo{person}{Lizi Liao}, \bibinfo{person}{Hao Fei}, \bibinfo{person}{Bobo Li}, {and} \bibinfo{person}{Jing Jiang}.} \bibinfo{year}{2024}\natexlab{}.
\newblock \showarticletitle{Actively Learn from LLMs with Uncertainty Propagation for Generalized Category Discovery}. In \bibinfo{booktitle}{\emph{North American Chapter of the Association for Computational Linguistics}}.
\newblock
\urldef\tempurl%
\url{https://aclanthology.org/2024.naacl-long.434.pdf}
\showURL{%
\tempurl}


\bibitem[Lin et~al\mbox{.}(2022)]%
        {Lin2022GenTUSSU}
\bibfield{author}{\bibinfo{person}{Hsien-Chin Lin}, \bibinfo{person}{Christian Geishauser}, \bibinfo{person}{Shutong Feng}, \bibinfo{person}{Nurul Lubis}, \bibinfo{person}{Carel van Niekerk}, \bibinfo{person}{Michael Heck}, {and} \bibinfo{person}{Milica Gavsi'c}.} \bibinfo{year}{2022}\natexlab{}.
\newblock \showarticletitle{GenTUS: Simulating User Behaviour and Language in Task-oriented Dialogues with Generative Transformers}.
\newblock \bibinfo{journal}{\emph{ArXiv}}  \bibinfo{volume}{abs/2208.10817} (\bibinfo{year}{2022}).
\newblock
\urldef\tempurl%
\url{https://aclanthology.org/2022.sigdial-1.28.pdf}
\showURL{%
\tempurl}


\bibitem[Lin et~al\mbox{.}(2021)]%
        {Lin2021DomainindependentUS}
\bibfield{author}{\bibinfo{person}{Hsien-Chin Lin}, \bibinfo{person}{Nurul Lubis}, \bibinfo{person}{Songbo Hu}, \bibinfo{person}{Carel van Niekerk}, \bibinfo{person}{Christian Geishauser}, \bibinfo{person}{Michael Heck}, \bibinfo{person}{Shutong Feng}, {and} \bibinfo{person}{Milica Gavsi'c}.} \bibinfo{year}{2021}\natexlab{}.
\newblock \showarticletitle{Domain-independent User Simulation with Transformers for Task-oriented Dialogue Systems}. In \bibinfo{booktitle}{\emph{SIGDIAL Conferences}}.
\newblock
\urldef\tempurl%
\url{https://aclanthology.org/2021.sigdial-1.47.pdf}
\showURL{%
\tempurl}


\bibitem[Liu et~al\mbox{.}(2021)]%
        {Liu2021TowardsES}
\bibfield{author}{\bibinfo{person}{Siyang Liu}, \bibinfo{person}{Chujie Zheng}, \bibinfo{person}{Orianna Demasi}, \bibinfo{person}{Sahand Sabour}, \bibinfo{person}{Yu Li}, \bibinfo{person}{Zhou Yu}, \bibinfo{person}{Yong Jiang}, {and} \bibinfo{person}{Minlie Huang}.} \bibinfo{year}{2021}\natexlab{}.
\newblock \showarticletitle{Towards Emotional Support Dialog Systems}. In \bibinfo{booktitle}{\emph{Annual Meeting of the Association for Computational Linguistics}}.
\newblock
\urldef\tempurl%
\url{https://aclanthology.org/2021.acl-long.269.pdf}
\showURL{%
\tempurl}


\bibitem[Lu et~al\mbox{.}(2024)]%
        {Lu2024LargeLM}
\bibfield{author}{\bibinfo{person}{Keming Lu}, \bibinfo{person}{Bowen Yu}, \bibinfo{person}{Chang Zhou}, {and} \bibinfo{person}{Jingren Zhou}.} \bibinfo{year}{2024}\natexlab{}.
\newblock \showarticletitle{Large Language Models are Superpositions of All Characters: Attaining Arbitrary Role-play via Self-Alignment}.
\newblock \bibinfo{journal}{\emph{ArXiv}}  \bibinfo{volume}{abs/2401.12474} (\bibinfo{year}{2024}).
\newblock
\urldef\tempurl%
\url{https://arxiv.org/pdf/2401.12474}
\showURL{%
\tempurl}


\bibitem[Men{\'e}ndez et~al\mbox{.}(1997)]%
        {menendez1997jensen}
\bibfield{author}{\bibinfo{person}{Mar{\'\i}a~Luisa Men{\'e}ndez}, \bibinfo{person}{JA Pardo}, \bibinfo{person}{L Pardo}, {and} \bibinfo{person}{MC Pardo}.} \bibinfo{year}{1997}\natexlab{}.
\newblock \showarticletitle{The jensen-shannon divergence}.
\newblock \bibinfo{journal}{\emph{Journal of the Franklin Institute}} \bibinfo{volume}{334}, \bibinfo{number}{2} (\bibinfo{year}{1997}), \bibinfo{pages}{307--318}.
\newblock


\bibitem[Moons and Vandervieren(2023)]%
        {Moons2023MeasuringAA}
\bibfield{author}{\bibinfo{person}{Filip Moons} {and} \bibinfo{person}{Ellen Vandervieren}.} \bibinfo{year}{2023}\natexlab{}.
\newblock \showarticletitle{Measuring agreement among several raters classifying subjects into one-or-more (hierarchical) nominal categories. A generalisation of Fleiss' kappa}.
\newblock
\urldef\tempurl%
\url{https://arxiv.org/pdf/2303.12502}
\showURL{%
\tempurl}


\bibitem[Peng et~al\mbox{.}(2018)]%
        {Peng2018DeepDI}
\bibfield{author}{\bibinfo{person}{Baolin Peng}, \bibinfo{person}{Xiujun Li}, \bibinfo{person}{Jianfeng Gao}, \bibinfo{person}{Jingjing Liu}, {and} \bibinfo{person}{Kam-Fai Wong}.} \bibinfo{year}{2018}\natexlab{}.
\newblock \showarticletitle{Deep Dyna-Q: Integrating Planning for Task-Completion Dialogue Policy Learning}. In \bibinfo{booktitle}{\emph{Annual Meeting of the Association for Computational Linguistics}}.
\newblock
\urldef\tempurl%
\url{https://aclanthology.org/P18-1203.pdf}
\showURL{%
\tempurl}


\bibitem[Revuz and Yor(1990)]%
        {Revuz1990ContinuousMA}
\bibfield{author}{\bibinfo{person}{Daniel Revuz} {and} \bibinfo{person}{Marc Yor}.} \bibinfo{year}{1990}\natexlab{}.
\newblock \showarticletitle{Continuous martingales and Brownian motion}.
\newblock
\urldef\tempurl%
\url{https://link.springer.com/book/10.1007/978-3-662-21726-9}
\showURL{%
\tempurl}


\bibitem[Scott and Bruce(1995)]%
        {Scott1995DecisionMakingST}
\bibfield{author}{\bibinfo{person}{Susanne~G. Scott} {and} \bibinfo{person}{Reginald~A. Bruce}.} \bibinfo{year}{1995}\natexlab{}.
\newblock \showarticletitle{Decision-Making Style: The Development and Assessment of a New Measure}.
\newblock \bibinfo{journal}{\emph{Educational and Psychological Measurement}}  \bibinfo{volume}{55} (\bibinfo{year}{1995}), \bibinfo{pages}{818 -- 831}.
\newblock
\urldef\tempurl%
\url{https://api.semanticscholar.org/CorpusID:143479230}
\showURL{%
\tempurl}


\bibitem[Sekuli'c et~al\mbox{.}(2024)]%
        {Sekulic2024ReliableLU}
\bibfield{author}{\bibinfo{person}{Ivan Sekuli'c}, \bibinfo{person}{Silvia Terragni}, \bibinfo{person}{Victor Guimaraes}, \bibinfo{person}{Nghia Khau}, \bibinfo{person}{Bruna Guedes}, \bibinfo{person}{Modestas Filipavicius}, \bibinfo{person}{Andre Manso}, {and} \bibinfo{person}{Roland Mathis}.} \bibinfo{year}{2024}\natexlab{}.
\newblock \showarticletitle{Reliable LLM-based User Simulator for Task-Oriented Dialogue Systems}.
\newblock \bibinfo{journal}{\emph{ArXiv}}  \bibinfo{volume}{abs/2402.13374} (\bibinfo{year}{2024}).
\newblock
\urldef\tempurl%
\url{https://aclanthology.org/2024.scichat-1.3.pdf}
\showURL{%
\tempurl}


\bibitem[Song et~al\mbox{.}(2020)]%
        {Song2020DenoisingDI}
\bibfield{author}{\bibinfo{person}{Jiaming Song}, \bibinfo{person}{Chenlin Meng}, {and} \bibinfo{person}{Stefano Ermon}.} \bibinfo{year}{2020}\natexlab{}.
\newblock \showarticletitle{Denoising Diffusion Implicit Models}.
\newblock \bibinfo{journal}{\emph{ArXiv}}  \bibinfo{volume}{abs/2010.02502} (\bibinfo{year}{2020}).
\newblock
\urldef\tempurl%
\url{https://arxiv.org/pdf/2010.02502}
\showURL{%
\tempurl}


\bibitem[Sutton et~al\mbox{.}(1999)]%
        {Sutton1999PolicyGM}
\bibfield{author}{\bibinfo{person}{Richard~S. Sutton}, \bibinfo{person}{David~A. McAllester}, \bibinfo{person}{Satinder Singh}, {and} \bibinfo{person}{Y. Mansour}.} \bibinfo{year}{1999}\natexlab{}.
\newblock \showarticletitle{Policy Gradient Methods for Reinforcement Learning with Function Approximation}. In \bibinfo{booktitle}{\emph{Neural Information Processing Systems}}.
\newblock
\urldef\tempurl%
\url{https://www.cis.upenn.edu/~mkearns/finread/Sutton.pdf}
\showURL{%
\tempurl}


\bibitem[Tavakoli et~al\mbox{.}(2022)]%
        {tavakoli2022analyzing}
\bibfield{author}{\bibinfo{person}{Leila Tavakoli}, \bibinfo{person}{Hamed Zamani}, \bibinfo{person}{Falk Scholer}, \bibinfo{person}{William~Bruce Croft}, {and} \bibinfo{person}{Mark Sanderson}.} \bibinfo{year}{2022}\natexlab{}.
\newblock \showarticletitle{Analyzing clarification in asynchronous information-seeking conversations}.
\newblock \bibinfo{journal}{\emph{Journal of the Association for Information Science and Technology}} \bibinfo{volume}{73}, \bibinfo{number}{3} (\bibinfo{year}{2022}), \bibinfo{pages}{449--471}.
\newblock


\bibitem[Tu et~al\mbox{.}(2024)]%
        {Tu2024CharacterEvalAC}
\bibfield{author}{\bibinfo{person}{Quan Tu}, \bibinfo{person}{Shilong Fan}, \bibinfo{person}{Zihang Tian}, {and} \bibinfo{person}{Rui Yan}.} \bibinfo{year}{2024}\natexlab{}.
\newblock \showarticletitle{CharacterEval: A Chinese Benchmark for Role-Playing Conversational Agent Evaluation}. In \bibinfo{booktitle}{\emph{Annual Meeting of the Association for Computational Linguistics}}.
\newblock
\urldef\tempurl%
\url{https://arxiv.org/pdf/2401.01275}
\showURL{%
\tempurl}


\bibitem[Ultes et~al\mbox{.}(2017)]%
        {Ultes2017PyDialAM}
\bibfield{author}{\bibinfo{person}{Stefan Ultes}, \bibinfo{person}{Lina~Maria Rojas-Barahona}, \bibinfo{person}{Pei hao Su}, \bibinfo{person}{David Vandyke}, \bibinfo{person}{Dongho Kim}, \bibinfo{person}{I{\~n}igo Casanueva}, \bibinfo{person}{Paweł Budzianowski}, \bibinfo{person}{Nikola Mrksic}, \bibinfo{person}{Tsung-Hsien Wen}, \bibinfo{person}{Milica Gavsic}, {and} \bibinfo{person}{Steve~J. Young}.} \bibinfo{year}{2017}\natexlab{}.
\newblock \showarticletitle{PyDial: A Multi-domain Statistical Dialogue System Toolkit}. In \bibinfo{booktitle}{\emph{Annual Meeting of the Association for Computational Linguistics}}.
\newblock
\urldef\tempurl%
\url{https://aclanthology.org/P17-4013.pdf}
\showURL{%
\tempurl}


\bibitem[Wang et~al\mbox{.}(2023a)]%
        {Wang2023DialoguePV}
\bibfield{author}{\bibinfo{person}{Jian Wang}, \bibinfo{person}{Dongding Lin}, {and} \bibinfo{person}{Wenjie Li}.} \bibinfo{year}{2023}\natexlab{a}.
\newblock \showarticletitle{Dialogue Planning via Brownian Bridge Stochastic Process for Goal-directed Proactive Dialogue}.
\newblock \bibinfo{journal}{\emph{ArXiv}}  \bibinfo{volume}{abs/2305.05290} (\bibinfo{year}{2023}).
\newblock
\urldef\tempurl%
\url{https://arxiv.org/pdf/2305.05290}
\showURL{%
\tempurl}


\bibitem[Wang et~al\mbox{.}(2019)]%
        {Wang2019PersuasionFG}
\bibfield{author}{\bibinfo{person}{Xuewei Wang}, \bibinfo{person}{Weiyan Shi}, \bibinfo{person}{Richard Kim}, \bibinfo{person}{Yoo~Jung Oh}, \bibinfo{person}{Sijia Yang}, \bibinfo{person}{Jingwen Zhang}, {and} \bibinfo{person}{Zhou Yu}.} \bibinfo{year}{2019}\natexlab{}.
\newblock \showarticletitle{Persuasion for Good: Towards a Personalized Persuasive Dialogue System for Social Good}.
\newblock \bibinfo{journal}{\emph{ArXiv}}  \bibinfo{volume}{abs/1906.06725} (\bibinfo{year}{2019}).
\newblock
\urldef\tempurl%
\url{https://aclanthology.org/P19-1566.pdf}
\showURL{%
\tempurl}


\bibitem[Wang et~al\mbox{.}(2023b)]%
        {Wang2023RoleLLMBE}
\bibfield{author}{\bibinfo{person}{Zekun~Moore Wang}, \bibinfo{person}{Zhongyuan Peng}, \bibinfo{person}{Haoran Que}, \bibinfo{person}{Jiaheng Liu}, \bibinfo{person}{Wangchunshu Zhou}, \bibinfo{person}{Yuhan Wu}, \bibinfo{person}{Hongcheng Guo}, \bibinfo{person}{Ruitong Gan}, \bibinfo{person}{Zehao Ni}, \bibinfo{person}{Man Zhang}, \bibinfo{person}{Zhaoxiang Zhang}, \bibinfo{person}{Wanli Ouyang}, \bibinfo{person}{Ke Xu}, \bibinfo{person}{Wenhu Chen}, \bibinfo{person}{Jie Fu}, {and} \bibinfo{person}{Junran Peng}.} \bibinfo{year}{2023}\natexlab{b}.
\newblock \showarticletitle{RoleLLM: Benchmarking, Eliciting, and Enhancing Role-Playing Abilities of Large Language Models}. In \bibinfo{booktitle}{\emph{Annual Meeting of the Association for Computational Linguistics}}.
\newblock
\urldef\tempurl%
\url{https://arxiv.org/pdf/2310.00746}
\showURL{%
\tempurl}


\bibitem[Wu et~al\mbox{.}(2025)]%
        {wu2025raiden}
\bibfield{author}{\bibinfo{person}{Bowen Wu}, \bibinfo{person}{Kaili Sun}, \bibinfo{person}{Ziwei Bai}, \bibinfo{person}{Ying Li}, {and} \bibinfo{person}{Baoxun Wang}.} \bibinfo{year}{2025}\natexlab{}.
\newblock \showarticletitle{RAIDEN Benchmark: Evaluating Role-playing Conversational Agents with Measurement-Driven Custom Dialogues}. In \bibinfo{booktitle}{\emph{Proceedings of the 31st International Conference on Computational Linguistics}}. \bibinfo{pages}{11086--11106}.
\newblock


\bibitem[Xiang et~al\mbox{.}(2024)]%
        {Xiang2024DiffusionDialogAD}
\bibfield{author}{\bibinfo{person}{Jianxiang Xiang}, \bibinfo{person}{Zhenhua Liu}, \bibinfo{person}{Haodong Liu}, \bibinfo{person}{Yin Bai}, \bibinfo{person}{Jia Cheng}, {and} \bibinfo{person}{Wenliang Chen}.} \bibinfo{year}{2024}\natexlab{}.
\newblock \showarticletitle{DiffusionDialog: A Diffusion Model for Diverse Dialog Generation with Latent Space}. In \bibinfo{booktitle}{\emph{International Conference on Language Resources and Evaluation}}.
\newblock
\urldef\tempurl%
\url{https://aclanthology.org/2024.lrec-main.440.pdf}
\showURL{%
\tempurl}


\bibitem[Ye et~al\mbox{.}(2024)]%
        {Ye2024SweetieChatAS}
\bibfield{author}{\bibinfo{person}{Jing Ye}, \bibinfo{person}{Lu Xiang}, \bibinfo{person}{Yaping Zhang}, {and} \bibinfo{person}{Chengqing Zong}.} \bibinfo{year}{2024}\natexlab{}.
\newblock \showarticletitle{SweetieChat: A Strategy-Enhanced Role-playing Framework for Diverse Scenarios Handling Emotional Support Agent}.
\newblock \bibinfo{journal}{\emph{ArXiv}}  \bibinfo{volume}{abs/2412.08389} (\bibinfo{year}{2024}).
\newblock
\urldef\tempurl%
\url{https://arxiv.org/pdf/2412.08389}
\showURL{%
\tempurl}


\bibitem[Yi et~al\mbox{.}(2018)]%
        {Yi2018ModelbasedRL}
\bibfield{author}{\bibinfo{person}{Fengji Yi}, \bibinfo{person}{Wenlong Fu}, {and} \bibinfo{person}{Huan Liang}.} \bibinfo{year}{2018}\natexlab{}.
\newblock \showarticletitle{Model-based reinforcement learning: A survey}.
\newblock
\urldef\tempurl%
\url{https://api.semanticscholar.org/CorpusID:108339287}
\showURL{%
\tempurl}


\bibitem[Yu et~al\mbox{.}(2023)]%
        {Yu2023PromptBasedMT}
\bibfield{author}{\bibinfo{person}{Xiao Yu}, \bibinfo{person}{Maximillian Chen}, {and} \bibinfo{person}{Zhou Yu}.} \bibinfo{year}{2023}\natexlab{}.
\newblock \showarticletitle{Prompt-Based Monte-Carlo Tree Search for Goal-Oriented Dialogue Policy Planning}. In \bibinfo{booktitle}{\emph{Conference on Empirical Methods in Natural Language Processing}}.
\newblock
\urldef\tempurl%
\url{https://arxiv.org/pdf/2305.13660}
\showURL{%
\tempurl}


\bibitem[Zhang et~al\mbox{.}(2023)]%
        {Zhang2023AskAE}
\bibfield{author}{\bibinfo{person}{Qiang Zhang}, \bibinfo{person}{Jason Naradowsky}, {and} \bibinfo{person}{Yusuke Miyao}.} \bibinfo{year}{2023}\natexlab{}.
\newblock \showarticletitle{Ask an Expert: Leveraging Language Models to Improve Strategic Reasoning in Goal-Oriented Dialogue Models}. In \bibinfo{booktitle}{\emph{Annual Meeting of the Association for Computational Linguistics}}.
\newblock
\urldef\tempurl%
\url{https://arxiv.org/pdf/2305.17878}
\showURL{%
\tempurl}


\bibitem[Zhang and Balog(2020)]%
        {Zhang2020EvaluatingCR}
\bibfield{author}{\bibinfo{person}{Shuo Zhang} {and} \bibinfo{person}{Krisztian Balog}.} \bibinfo{year}{2020}\natexlab{}.
\newblock \showarticletitle{Evaluating Conversational Recommender Systems via User Simulation}.
\newblock \bibinfo{journal}{\emph{Proceedings of the 26th ACM SIGKDD International Conference on Knowledge Discovery \& Data Mining}} (\bibinfo{year}{2020}).
\newblock
\urldef\tempurl%
\url{https://dl.acm.org/doi/10.1145/3394486.3403202}
\showURL{%
\tempurl}


\bibitem[Zhang et~al\mbox{.}(2024)]%
        {Zhang2024StrengthLI}
\bibfield{author}{\bibinfo{person}{Tong Zhang}, \bibinfo{person}{Chen Huang}, \bibinfo{person}{Yang Deng}, \bibinfo{person}{Hongru Liang}, \bibinfo{person}{Jia Liu}, \bibinfo{person}{Zujie Wen}, \bibinfo{person}{Wenqiang Lei}, {and} \bibinfo{person}{Tat-Seng Chua}.} \bibinfo{year}{2024}\natexlab{}.
\newblock \showarticletitle{Strength Lies in Differences! Improving Strategy Planning for Non-collaborative Dialogues via Diversified User Simulation}. In \bibinfo{booktitle}{\emph{Conference on Empirical Methods in Natural Language Processing}}.
\newblock
\urldef\tempurl%
\url{https://aclanthology.org/2024.emnlp-main.26.pdf}
\showURL{%
\tempurl}


\bibitem[Zhang et~al\mbox{.}(2019)]%
        {Zhang2019BudgetedPL}
\bibfield{author}{\bibinfo{person}{Zhirui Zhang}, \bibinfo{person}{Xiujun Li}, \bibinfo{person}{Jianfeng Gao}, {and} \bibinfo{person}{Enhong Chen}.} \bibinfo{year}{2019}\natexlab{}.
\newblock \showarticletitle{Budgeted Policy Learning for Task-Oriented Dialogue Systems}. In \bibinfo{booktitle}{\emph{Annual Meeting of the Association for Computational Linguistics}}.
\newblock
\urldef\tempurl%
\url{https://aclanthology.org/P19-1364.pdf}
\showURL{%
\tempurl}


\bibitem[Zhao et~al\mbox{.}(2023)]%
        {Zhao2023IsCE}
\bibfield{author}{\bibinfo{person}{Weixiang Zhao}, \bibinfo{person}{Yanyan Zhao}, \bibinfo{person}{Xin Lu}, \bibinfo{person}{Shilong Wang}, \bibinfo{person}{Yanpeng Tong}, {and} \bibinfo{person}{Bing Qin}.} \bibinfo{year}{2023}\natexlab{}.
\newblock \showarticletitle{Is ChatGPT Equipped with Emotional Dialogue Capabilities?}
\newblock \bibinfo{journal}{\emph{ArXiv}}  \bibinfo{volume}{abs/2304.09582} (\bibinfo{year}{2023}).
\newblock
\urldef\tempurl%
\url{https://arxiv.org/pdf/2304.09582}
\showURL{%
\tempurl}


\end{thebibliography}

\end{document}